%% file: main.tex
\documentclass[conference]{IEEEtran}
\IEEEoverridecommandlockouts

\usepackage{cite}
\usepackage{amsmath,amsopn,amssymb}
\usepackage{graphicx,xspace,color,soul}
\usepackage{epsfig,subfigure}
\usepackage{longtable,multirow}
\usepackage{array,float}
\usepackage{url}
\usepackage{algorithm,algorithmicx,algpseudocode}
\usepackage{bm,epstopdf,booktabs}

\renewcommand{\vec}[1]{\mathbf{#1}}

\begin{document}

\title{Exploring Structure-Adaptive Graph Learning for Robust Semi-Supervised Classification}

\author{Xiang Gao, Wei Hu, Zongming Guo\\
Wangxuan Institute of Computer Technology, Peking University, Beijing, China\\
{\tt\small \{gyshgx868, forhuwei, guozongming\}@pku.edu.cn}
}

\maketitle

\begin{abstract}

Graph Convolutional Neural Networks (GCNNs) are generalizations of CNNs to graph-structured data, in which convolution is guided by the graph topology. 
In many cases where graphs are unavailable, existing methods manually construct graphs or learn task-driven adaptive graphs. 
In this paper, we propose Graph Learning Neural Networks (GLNNs), which exploit the optimization of graphs (the adjacency matrix in particular) from both data and tasks. 
Leveraging on spectral graph theory, we propose the objective of graph learning from a sparsity constraint, properties of a valid adjacency matrix as well as a graph Laplacian regularizer via maximum a posteriori estimation. 
The optimization objective is then integrated into the loss function of the GCNN, which adapts the graph topology to not only labels of a specific task but also the input data.
Experimental results show that our proposed GLNN outperforms state-of-the-art approaches over widely adopted social network datasets and citation network datasets for semi-supervised classification.
\end{abstract}

\section{Introduction}
\label{sec:intro}
\input{01_introduction.tex}

\section{Related Work}
\label{sec:related}

\input{02_related_work.tex}

\section{Background in Spectral Graph Theory}
\label{sec:Laplacian}
\input{Laplacian.tex}

\section{The Proposed Graph Learning}
\label{sec:graph}
\input{03_graph_learning.tex}

\section{The Proposed GLNN}
\label{sec:method}
\input{04_method.tex}

\section{Experimental Results}
\label{sec:results}
\input{05_results.tex}

\section{Conclusion}
\label{sec:conclude}
\input{06_conclusion.tex}

\end{document}

%% file: 01_introduction.tex
Graphs are natural and efficient representation for non-Euclidean data, such as social networks, citation networks, brain neural networks, 3D geometric data, etc. Nodes of a graph often represent entities and edges reflect the relationship between two nodes. 
In various applications of graph data, semi-supervised classification is crucial due to the high cost of labeling, where labels are only available for a small subset of nodes.
This enables a variety of applications, such as suggesting new connections to individuals based on similar interests or experiences, sociological study of communities\footnote{For instance, the extent to which communities form around particular interests or affiliations.}, and detecting trends in academic research from a citation network \cite{bhagat2011node,asatani2018Detecting}.

While semi-supervised learning has been studied extensively for Euclidean data, its extension to non-Euclidean domain is challenging due to data irregularity. Recently, Graph Convolutional Neural Networks (GCNNs) have been proposed to generalize CNNs to non-Euclidean domain \cite{bruna2013spectral}, which has shown its efficiency in various applications including semi-supervised classification \cite{kipf2016semi,defferrard2016convolutional,gori2005new,duvenaud2015convolutional,bruna2013spectral,niepert2016learning}. 
Graphs are fed into the network for basic operations such as graph convolution, and thus play a vital role in feature extraction. 
In cases where graphs are not readily available, many previous works manually construct graphs from the input data, which tend to be sub-optimal. 

\begin{figure}[t]
    \centering
    \includegraphics[width=8.3cm]{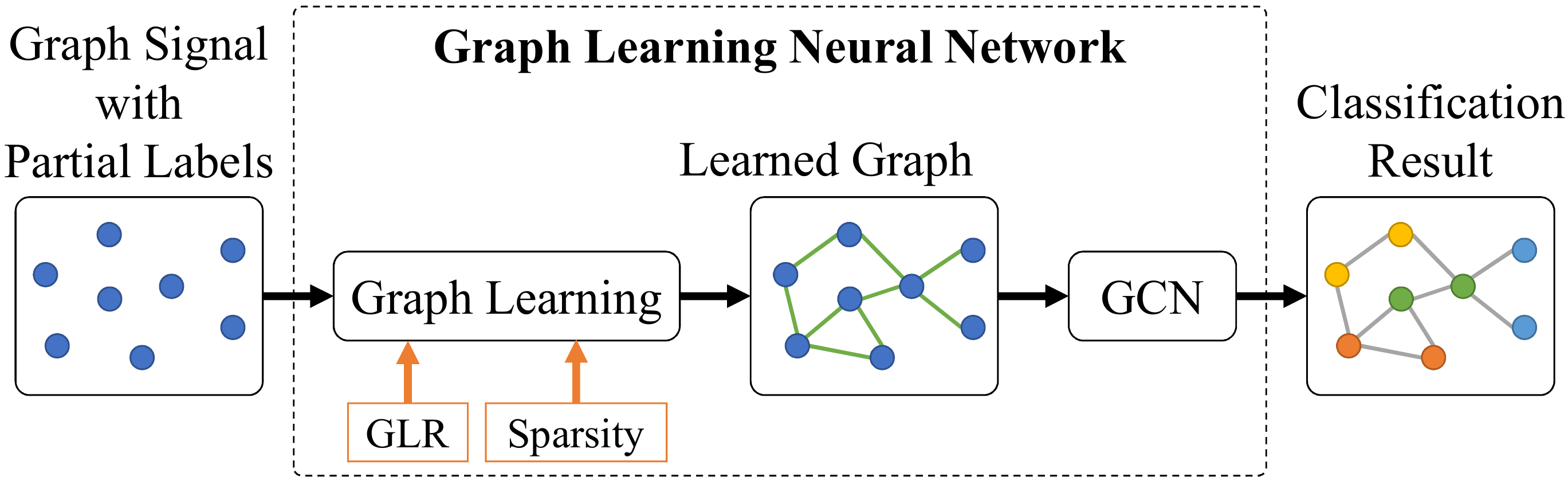}
    \caption{\textbf{The pipeline of the proposed GLNN for semi-supervised node classification.} Our proposed GLNN takes graph signals as the input, which first goes through a graph learning layer and outputs a learned graph (green lines indicate learned edges). We then feed the learned graph into two-layer graph convolution as in the GCN along with the graph signals for feature learning, and finally output the classification results.}
    \label{fig:teaser}
\end{figure}

Recently, there have been few attempts at graph learning in GCNNs.
Li et al. \cite{li2018adaptive} learn an adaptive graph via distance metric learning for graph classification, which is task-driven during the model optimization.   
Jiang et al. \cite{jiang19cvpr} propose a graph learning convolutional network, which learns a non-negative edge weight function that represents pairwise similarities within graph data for semi-supervised learning. 
However, only edge weights are learned while edge connectivities resort to the structure of the given ground truth graph, which may not capture potential connections.
Li et al. \cite{li2019spatio} propose a spatio-temporal graph learning scheme for skeleton-based action recognition, which adaptively learns intrinsic high-order connectivities for skeleton joints.
Shi et al. \cite{shi2019two} come up with a two-stream adaptive graph convolutional network for skeleton-based action recognition, where a shared graph for all instances and an individual graph for each instance are learned in an end-to-end manner.
Nevertheless, the optimization of the above network models either are task-driven or focus on edge weight learning based on an available graph structure.

Extending previous works on graph learning for GCNNs, we propose a structure-adaptive  Graph Learning Neural Network (GLNN) for \textit{robust} semi-supervised node classification, as shown in Fig.~\ref{fig:teaser}. 
The key idea is to pose graph learning as an optimization problem of an adjacency matrix\footnote{The adjacency matrix encodes connectivities and edge weights of a graph.} from both data and labels based on a structure-adaptive  regularization---\textit{Graph Laplacian Regularizer} (GLR) \cite{Shuman2013The}. 
GLR measures the smoothness of data with respect to the graph Laplacian matrix\footnote{In spectral graph theory, a graph Laplacian matrix is an algebraic representation of connectivities and node degrees of the corresponding graph, which can be computed from the adjacency matrix and will be introduced in Section III.}, which essentially enforces the graph to capture pairwise similarities within the data.  
This structure-adaptive regularization enhances the robustness of the network model especially in case of low label rates. 

Specifically, we first represent features of network data (e.g., social/citation networks) as signals on the graph, and pose a Maximum a Posteriori (MAP) estimation on the underlying adjacency matrix. In the MAP estimation, we propose a likelihood function based on GLR. Also, we propose a prior distribution from the sparsity constraint and properties of a valid adjacency matrix, considering a symmetric, normalized and loopless graph. The MAP estimation then leads to an optimization problem for the underlying adjacency matrix. 
Secondly, we integrate the optimization objective into the cross-entropy loss function of a GCNN, thus optimizing the network model in both \textbf{structure-adaptive}  and \textbf{task-driven} manners. 
Finally, we design the framework of the proposed GLNN, which mainly consists of a Graph Learning Layer and Graph Convolution Layers. 

In summary, our main contributions are as follows:

\begin{itemize}
    \item We propose graph learning regularized by GLR for robust semi-supervised node classification especially at low label rates, which jointly learns graph connectivities and edge weights in both structure-adaptive  and task-driven manners. 
    \item We present a GLNN framework that integrates graph learning with graph convolution, which optimizes the network model by introducing GLR, properties of valid adjacency matrices and sparsity constraints into the loss function. 
    \item Extensive experiments demonstrate the superiority and robustness of our method compared to state-of-the-art approaches on widely used citation network datasets and social network datasets. 
\end{itemize}


%% file: 02_related_work.tex
Quite a few approaches for semi-supervised node classification have been proposed in recent years, most of which fall into three categories: label-smoothness based methods, graph embedding based methods, and convolutional neural network methods.

\textbf{Label-smoothness based methods.} This class of methods are based on the assumption that neighboring nodes tend to have the same labels. Belkin et al. \cite{belkin2006manifold} deploy a parameterized classifier in Reproducing Kernel Hilbert Space (RKHS), which is inductive and naturally handles unobserved instances. Weston et al. \cite{weston2012deep} propose a deep semi-supervised embedding approach, which extends the regularization and imposes stronger constraints on a neural network. 

\textbf{Graph embedding based methods.} This class of approaches are inspired by the skip-gram model \cite{mikolov2013distributed}. Perozzi et al. \cite{perozzi2014deepwalk} propose a DeepWalk model, which learns embeddings via the prediction of the local neighborhood of nodes, sampled from random walks on the graph. Yang et al. \cite{yang2016revisiting} alleviate the optimization step of DeepWalk by injecting label information in the process of learning embeddings. Tran et al. \cite{tran2018learning} present a novel autoencoder architecture (LoNGAE) capable of learning a joint representation of both local graph structure and available node features for the multitask learning of link prediction and node classification. 

\textbf{Convolutional neural network methods.} This family of approaches typically employ CNNs, GCNs, or their variants. Kipf et al. \cite{kipf2016semi} propose a simplified graph convolution model for semi-supervised learning by employing a first-order approximation of spectral filters. Monti et al. \cite{monti2017geometric} propose mixture model (MoNet) and provide a unified generalization of CNN architectures on graphs. Velickovic et al. \cite{velivckovic2018graph} present Graph Attention Networks (GATs) for semi-supervised learning by designing an attention layer. Xu et al. \cite{xu2018graph} present graph wavelet neural network (GWNN) to implement efficient convolution on graph data, which takes graph wavelets instead of the eigenvectors of the graph Laplacian as a set of basis.

%% file: Laplacian.tex
\subsection{Graph Laplacian}

We consider an undirected graph $ \mathcal{G}=\{\mathcal{V},\mathcal{E},\mathbf{A}\} $ composed of a node set $ \mathcal{V} $ of cardinality $|\mathcal{V}|=N$, an edge set $\mathcal{E}$ connecting nodes, and a weighted \textit{adjacency matrix} $\mathbf{A}$. $\mathbf{A}$ is a real symmetric $N \times N$ matrix, where $a_{i,j}$ is the weight assigned to the edge $(i,j)$ connecting nodes $i$ and $j$.

The \textit{graph Laplacian} is defined from the adjacency matrix. 
Among different variants of Laplacian matrices, the commonly used \textit{combinatorial graph Laplacian} \cite{chung1997spectral} is defined as 
\begin{equation}
   \mathbf{L}=\mathbf{D}-\mathbf{A}, 
   \label{eq:laplacian}
\end{equation}
where $ \mathbf{D} $ is the \textit{degree matrix}---a diagonal matrix where $ d_{i,i} = \sum_{j=1}^N a_{i,j} $.

Considering the numerical stability in a deep neural network model, we employ the symmetric \textit{normalized} Laplacian \cite{chung1997spectral}, which is defined as
\begin{equation}
    \mathcal{L} =\mathbf{D}^{-\frac{1}{2}}\mathbf{L}\mathbf{D}^{-\frac{1}{2}}=\mathbf{I}-\mathbf{D}^{-\frac{1}{2}}\mathbf{A}\mathbf{D}^{-\frac{1}{2}},
    \label{eq:normailized_laplacian}
\end{equation}
where $\mathbf{I}$ is an identity matrix.

\subsection{Graph Laplacian Regularizer}

\textit{Graph signal} refers to data that resides on the nodes of a graph. 
In our context, we treat each person or paper as a node in a graph, and the relationship as the edge between each pair of nodes. 
Then we define the corresponding graph signal as the social feature or paper information of each node.

A graph signal $ \vec{x} \in \mathbb{R}^N $ defined on a graph $ \mathcal{G} $ is \textit{smooth} with respect to $ \mathcal{G} $ \cite{Shuman2013The} if  
\begin{equation}
	\vec{x}^{\top} \mathbf{L} \vec{x} =\sum_{i=1}^{N} \sum_{j=1}^{N} w_{i,j}(x_i - x_j)^2 < \epsilon,
	\label{eq:glr}
\end{equation}
where $ \epsilon $ is a small positive scalar. 
To satisfy (\ref{eq:glr}), disconnected or weakly connected node pair $ x_i $ and $ x_j $ with significantly different values must correspond to a small edge weight $ a_{i,j} $; 
for strongly connected $x_i$ and $x_j$ with similar values, $ a_{i,j} $ can be large.  
Hence, (\ref{eq:glr}) forces $ \mathcal{G} $ to capture pairwise similarities in $ \vec{x} $, and is commonly called the \textit{graph Laplacian Regularizer} (GLR) \cite{Shuman2013The}.

%% file: 03_graph_learning.tex
In this section, we elaborate on the problem formulation of structure-adaptive graph learning, and then describe the corresponding network model optimization.

\subsection{Problem Formulation of Graph Learning}
We propose to exploit the learning of the direct representation of the underlying graph, i.e., the adjacency matrix. 
In particular, we pose a MAP estimation of the adjacency matrix, which leads to graph learning from the GLR, sparsity constraint and properties of a valid adjacency matrix.

\subsubsection{MAP Estimation of Graph Adjacency Matrices}

We first formulate a MAP estimation problem for the optimal graph adjacency matrix $\mathbf{\hat{A}}$: given the observed graph signal $\vec{x}$, find the most probable graph $\mathbf{\hat{A}}$,
\begin{equation}
    \vec{\tilde{A}}_\text{MAP}(\mathbf{x}) = \arg \max_{\mathbf{\hat{A}}} \ f(\vec{x} \mid \mathbf{\hat{A}})g(\mathbf{\hat{A}}),
    \label{eq:final_map}
\end{equation}
where $f(\vec{x} \mid \mathbf{\hat{A}})$ is the likelihood function, and $g(\mathbf{\hat{A}})$ is the prior probability distribution of $\mathbf{\hat{A}}$. The likelihood function $f(\vec{x} \mid \mathbf{\hat{A}})$ is the probability of obtaining the observed graph signal $\vec{x}$ given the graph $\mathbf{\hat{A}}$, while the prior probability distribution $g(\mathbf{\hat{A}})$ provides the prior knowledge of $\mathbf{\hat{A}}$. 
The likelihood and prior functions are discussed in order as follows.

\begin{figure*}[t]
    \centering
    \includegraphics[width=\textwidth]{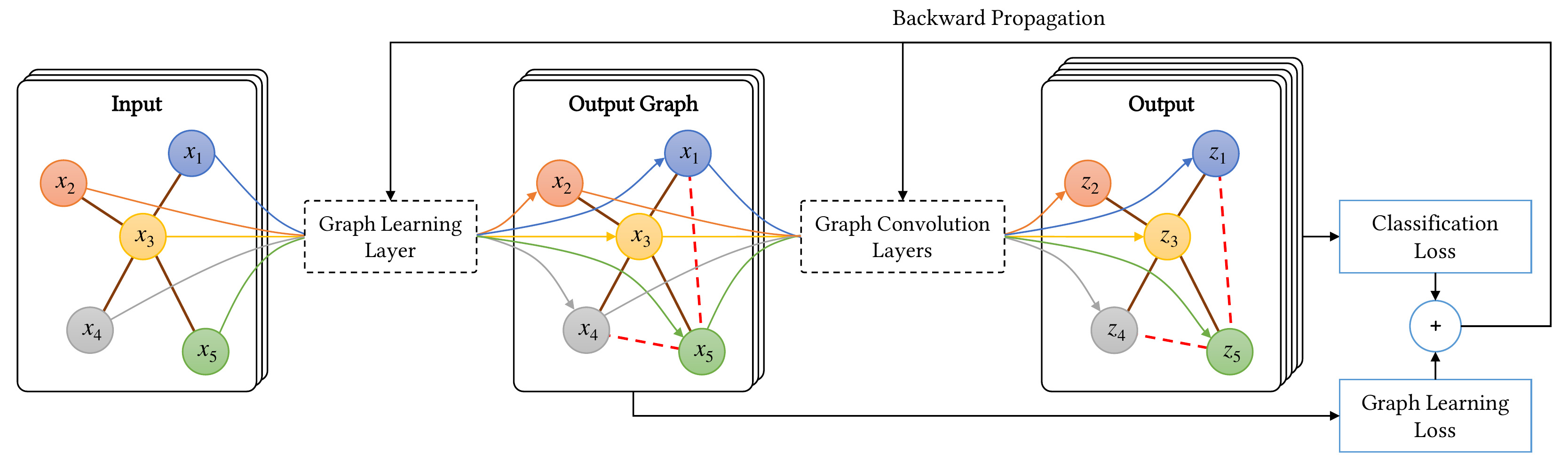}
    \caption{\textbf{The architecture of the proposed GLNN for semi-supervised node classification.} Our proposed network takes a feature map and a ground-truth graph (optional) as the input in the training phase. We then employ the Graph Learning Layer to obtain an optimal graph, and feed the learned graph and feature map to Graph Convolution Layers to generate the output node classification scores. We add classification loss and the proposed graph learning loss together for backward propagation.}
    \label{fig:pipeline}
\end{figure*}

\subsubsection{Proposed Prior Probability Distribution}
The prior probability distribution $g(\mathbf{\hat{A}})$ provides the prior knowledge of $\mathbf{\hat{A}}$. We propose the prior knowledge of an adjacency matrix from two aspects: 1) the sparsity constraint $g_s(\mathbf{\hat{A}})$; 2) the property constraint of a valid adjacency matrix $g_p(\mathbf{\hat{A}})$. Since the two prior constraints are independent, we have
\begin{equation}
    g(\mathbf{\hat{A}}) = g_s(\mathbf{\hat{A}})g_p(\mathbf{\hat{A}}).
    \label{eq:prior_distribution}
\end{equation}
We elaborate on the sparsity constraint and property constraint separately as follows.

\textbf{Sparsity Constraint.}
We pose a sparsity constraint on $\mathbf{\hat{A}}$. 
This is based on the observation that the similarity relationship in real-world data is often sparse. 
Also, this constraint avoids data overfitting by capturing prominent similarities.  
Mathematically, we define the sparsity prior as
\begin{equation}
    g_s(\mathbf{\hat{A}}) = \exp \left( -\lambda_1 \| \mathbf{\hat{A}} \|_1 \right),
    \label{eq:sparsity_constraint}
\end{equation}
where $\| \cdot \|_1$ denotes the $l_1$-norm of a matrix, and $\lambda_1$ is a weighting parameter. 

\textbf{Property Constraint.}
The properties of a valid normalized adjacency matrix include:
\begin{itemize}
    \item Symmetry. In our model, we only consider undirected graphs, in which any edge $(x_i, x_j) \in \mathcal{E}$ is identical to the edge $(x_j, x_i) \in \mathcal{E}$. Hence, the corresponding adjacency matrix $\mathbf{\hat{A}}$ is symmetric. The mathematical description is
    \begin{equation}
        \mathbf{\hat{A}}^\top = \mathbf{\hat{A}}.
    \end{equation}
    \item Normalized. We consider normalized adjacency matrices in order to avoid numerical instabilities and exploding or vanishing gradients when used in a deep neural network architecture. The mathematical description is
    \begin{equation}
        \mathbf{\hat{A}}\mathbf{1} = \mathbf{1},
    \end{equation}
    where $\vec{1}$ denotes the vector with all the elements equal to one.
    \item Loopless. Since it is often unnecessary to link a node with itself such as in a citation network or social network, we consider graphs without self loop. The mathematical description is
    \begin{equation}
        \text{tr}(\mathbf{\hat{A}}) = 0,
    \end{equation}
    where $\text{tr}(\cdot)$ denotes the trace of a matrix.
\end{itemize}

Combining the above properties, we propose the following prior probabilistic distribution of $\mathbf{\hat{A}}$:   
\begin{equation}
\begin{aligned}
    g_p(\mathbf{\hat{A}}) = \exp \left( \right. & -\lambda_2 \| \mathbf{\hat{A}}^\top - \mathbf{\hat{A}} \|_F^2 \\
    & \left.-\lambda_3 \| \mathbf{\hat{A}}\mathbf{1} - \mathbf{1} \|_F^2 -\lambda_4 | \text{tr}(\mathbf{\hat{A}}) |^2 \right),
\end{aligned}
\label{eq:properties_constraints}
\end{equation}
where $\| \cdot \|_F$ is the Frobenius norm of a matrix, and $\lambda_2$, $\lambda_3$, and $\lambda_4$ are all weighting parameters.

\subsubsection{Proposed Likelihood Function}

Since GLR enforces the graph signal $\vec{x}$ to adapt to the topology of the graph described by $\mathbf{\hat{A}}$, we propose the likelihood function as 
\begin{equation}
\begin{split}
    f(\vec{x} \mid \mathbf{\hat{A}}) 
    & = \text{exp} \left(-\lambda_0 \vec{x}^\top \mathbf{L} \vec{x} \right)\\
    & = \text{exp} \left(-\lambda_0 \vec{x}^\top ( \mathbf{I} - \mathbf{\hat{A}} ) \vec{x} \right),
    \label{eq:final_likelihood_func}
\end{split}
\end{equation}
where $\lambda_0$ is a parameter. 
Since we assume $\mathbf{\hat{A}}$ is normalized, the degree matrix is an identity matrix as in \eqref{eq:final_likelihood_func}.  
This provides a structural assumption of data encoded in the adjacency matrix, which includes both graph connectivities and edge weights.

\subsubsection{Final MAP Estimation} 
Having discussed the proposed likelihood function and prior, we arrive at our final MAP estimation of the graph adjacency matrix. Combining (\ref{eq:final_map}),  (\ref{eq:prior_distribution}), (\ref{eq:sparsity_constraint}), (\ref{eq:properties_constraints}) and (\ref{eq:final_likelihood_func}), we have

\begin{equation}
\begin{split}
    \max_{\mathbf{\hat{A}}} \ \ & \exp\left(-\lambda_0 \vec{x}^\top(\mathbf{I} - \mathbf{\hat{A}}) \vec{x} \right) \cdot \exp\left( -\lambda_1 \| \mathbf{\hat{A}} \|_1 \right) \\
    & \begin{aligned}
    \cdot \exp \left( \right. & -\lambda_2 \| \mathbf{\hat{A}}^\top - \mathbf{\hat{A}} \|_F^2 -\lambda_3 \| \mathbf{\hat{A}}\mathbf{1} - \mathbf{1} \|_F^2 \\
    & \left. -\lambda_4 | \text{tr}(\mathbf{\hat{A}}) |^2 \right).
    \end{aligned} 
    \label{eq:formulation}
\end{split}
\end{equation}

Taking the logarithm of (\ref{eq:formulation}) and multiplying by $-1$, we have
\begin{equation}
\begin{split}
    \min_{\mathbf{\hat{A}}} \ \ & \lambda_0 \vec{x}^\top(\mathbf{I} - \mathbf{\hat{A}}) \vec{x} + \lambda_1 \| \mathbf{\hat{A}} \|_1 + \lambda_2 \| \mathbf{\hat{A}}^\top - \mathbf{\hat{A}} \|_F^2 \\
    & + \lambda_3 \| \mathbf{\hat{A}}\mathbf{1} - \mathbf{1} \|_F^2 + \lambda_4 | \text{tr}(\mathbf{\hat{A}}) |^2,
    \label{eq:final_formulation}
\end{split}
\end{equation}
where the first term is GLR, the second term is the sparsity constraint of the adjacency matrix, and the rest are the property constraints of a valid adjacency matrix. Hence, the above problem formulation is to learn such a valid and sparse adjacency matrix $\mathbf{\hat{A}}$ that the graph signal $\mathbf{\vec{x}}$ is adaptive to the learned graph. 

\subsection{Network Model Optimization}
We propose to integrate the optimization objective in (\ref{eq:final_formulation}) into the loss function of the GCN, which seamlessly optimizes the network model with graph learning. Accordingly, the proposed overall loss in graph learning $\mathcal{L}_\text{GL}$ includes three components: the loss in GLR $\mathcal{L}_\text{GLR}$, the loss in sparsity $\mathcal{L}_\text{sparsity}$, and the loss in property constraints $\mathcal{L}_\text{properties}$:
\begin{equation}
    \mathcal{L}_\text{GL} = \mathcal{L}_\text{GLR} +  \mathcal{L}_\text{sparsity} + \mathcal{L}_\text{properties}.
    \label{eq:loss_gl}
\end{equation}

In accordance with (\ref{eq:final_formulation}),  $\mathcal{L}_\text{GLR}$ is defined as
\begin{equation}
    \mathcal{L}_\text{GLR} = \lambda_0 \| \mathbf{x}^\top (\mathbf{I} - \mathbf{\hat{A}}_\text{out}) \mathbf{x} \|_2^2.
\end{equation}
where $\mathbf{\hat{A}}_\text{out}$ is the final output adjacency matrix at each epoch. 
$\mathcal{L}_\text{GLR}$ is in a quadratic form, which is differentiable.

$\mathcal{L}_\text{sparsity}$ is accordingly defined as
\begin{equation}
    \mathcal{L}_\text{sparsity} = \lambda_1 \| \mathbf{\hat{A}}_\text{out} \|_1.
\end{equation}
The $l_1$-norm is non-differentiable since it is not continuous everywhere. In the backward propagation algorithm of machine learning, the gradient of the $l_1$ regularization is solved using the sub-gradient \cite{sra2012optimization}:
\begin{equation}
    \frac{\partial \mathcal{L}_\text{sparsity}}{\partial \mathbf{\hat{A}}_\text{out}} = \text{sgn}(\mathbf{\hat{A}}_\text{out}),
\end{equation}
where
\begin{equation}
    \text{sgn}(x)=\left\{
\begin{array}{lcl}
-1       &      & x < 0, \\
0     &      & x = 0, \\
1     &      & x > 0.
\end{array} \right.
\end{equation}

Finally, we define $\mathcal{L}_\text{properties}$ as
\begin{equation}
\begin{split}
    \mathcal{L}_\text{properties} = & \lambda_2 \| \mathbf{\hat{A}}_\text{out}^\top - \mathbf{\hat{A}}_\text{out} \|_2^2 \\
    & + \lambda_3 \| \mathbf{\hat{A}}_\text{out} \mathbf{1} - \mathbf{1} \|_2^2 + \lambda_4 | \text{tr}(\mathbf{\hat{A}}_\text{out}) |^2.
\end{split}
\label{eq:loss_properties}
\end{equation}
The three terms of this function are all $l_2$ norms, and thus the function is differentiable. 

Hence, we are able to compute the gradient of the proposed loss function for graph learning, which is suitable to guide the optimization of the network model.

%% file: 04_method.tex



Having discussed the proposed graph learning, we elaborate on its integration with GCN, which leads to the proposed GLNN architecture. As shown in Fig.~\ref{fig:pipeline}, given graph-structured networks as the input, GLNN consists of two major procedures: graph learning and graph convolution. The graph learning layer aims to produce an optimal adjacency matrix for the subsequent graph convolution. The graph convolution has two layers, with the same parameter settings as in \cite{kipf2016semi}. Besides, the classification loss measures the classification performance of our model, whose value is a probability between 0 and 1. The graph learning loss is for the optimization of the adjacency matrix during the network training. We train the entire network using gradient descent via the backward propagation of both the classification loss and graph learning loss. We discuss each module in detail as follows.

\subsection{Graph Learning Layer}

The graph learning layer aims to construct an optimal graph structure from the input graph signals. In this layer, we first randomly initialize the adjacency matrix before training the network, and then train the graph using gradient descent from the backward propagation of the proposed loss function introduced in (\ref{eq:loss_gl}).

In order to further simplify the implementation, we perform the following symmetrization on the adjacency matrix $\mathbf{\hat{A}}$ before the output of the graph learning layer, which aims to ensure the symmetry property:
\begin{equation}
    \mathbf{\hat{A}}_\text{out} = \frac{1}{2} \left(\mathbf{\hat{A}}^\top + \mathbf{\hat{A}} \right),
    \label{eq:symmetrization}
\end{equation}
where $\mathbf{\hat{A}}_\text{out} \in \mathbb{R}^{N \times N}$ is the final output adjacency matrix of the graph learning layer at each epoch. It is a real symmetric matrix by definition. Thus, we can remove the symmetric term in (\ref{eq:loss_properties}).

The graph obtained from the graph learning layer is then fed into the GCN architecture for the subsequent semi-supervised node classification task.

\subsection{Graph Convolution and Feature Transfer }

As introduced in \cite{kipf2016semi}, the graph convolution to a signal $\mathbf{X} \in \mathbb{R}^{N \times C}$ with $C$ input channels (i.e., a $C$-dimensional feature vector for every node) and $F$ filters are defined as follows:
\begin{equation}
    \mathbf{Z}=\sigma \left( \mathbf{\hat{A}}_\text{out} \mathbf{X}\mathbf{W} \right),
\end{equation}
where $\mathbf{W} \in \mathbb{R}^{C \times F}$ is a matrix of filter parameters, $\mathbf{Z} \in \mathbb{R}^{N \times F}$ is the convoluted signal matrix, and $\sigma(\cdot)$ is an activation function.

We consider a two-layer GCN with the learned adjacency matrix $\mathbf{\hat{A}}_\text{out}$. The forward model takes the form:
\begin{equation}
\begin{split}
    \mathbf{Z} &= f(\mathbf{X}, \mathbf{\hat{A}}_\text{out}) \\
    &= \text{softmax}\left(\mathbf{\hat{A}}_\text{out} \ \text{ReLU}\left(\mathbf{\hat{A}}_\text{out}\mathbf{X}\mathbf{W}^{(0)}\right)\mathbf{W}^{(1)}\right),
\end{split}
\end{equation}
where $\mathbf{W}^{(0)} \in \mathbb{R}^{C \times H}$ is an input-to-hidden weight matrix for a hidden graph convolutional layer with $H$ output feature channels, and $\mathbf{W}^{(1)} \in \mathbb{R}^{H \times F}$ is a hidden-to-output weight matrix for an output layer with $F$ classes. The $\text{softmax}(\cdot)$ and $\text{ReLU}(\cdot)$ are two activation functions. 

\subsection{The Overall Loss Function}
In semi-supervised node classification tasks, we need to evaluate the cross-entropy loss (the Classification Loss in Fig.~\ref{fig:pipeline}) over the labeled data:
\begin{equation}
    \mathcal{L}_\text{GCN} = - \sum_{l \in \mathcal{Y}_L} \sum_{f=1}^{F} \mathbf{Y}_{lf} \ln{\mathbf{Z}_{lf}},
\end{equation}
where $\mathcal{Y}_L$ is a set of labeled nodes, $\mathbf{Y}_{l \cdot}$ denotes the label for the $l^\text{th}$ labeled node, and $\mathbf{Z}_{l \cdot}$ is the predicted label for the $l^\text{th}$ node.

Further, in cases where the ground truth graph of training samples is available, we add the following loss function to ensure the learned graph is closer to the ground truth:
\begin{equation}
    \mathcal{L}_\text{gt} = \| \mathbf{\hat{A}}_\text{out} - \mathbf{A}_\text{gt} \|_2^2,
\end{equation}
where $\mathbf{A}_\text{gt}$ is a ground truth adjacency matrix.

Hence, the overall loss function of our proposed GLNN framework is
\begin{equation}
    \mathcal{L}_\text{overall} = \mathcal{L}_\text{GCN} + \mathcal{L}_\text{GL} + \alpha \mathcal{L}_\text{gt}.
    \label{eq:overall_loss}
\end{equation}
where $\alpha$ is a weighting parameter.

\begin{figure*}[htbp]
    \centering
    \subfigure[TerroristsRel: GT]{\includegraphics[width=0.19\textwidth]{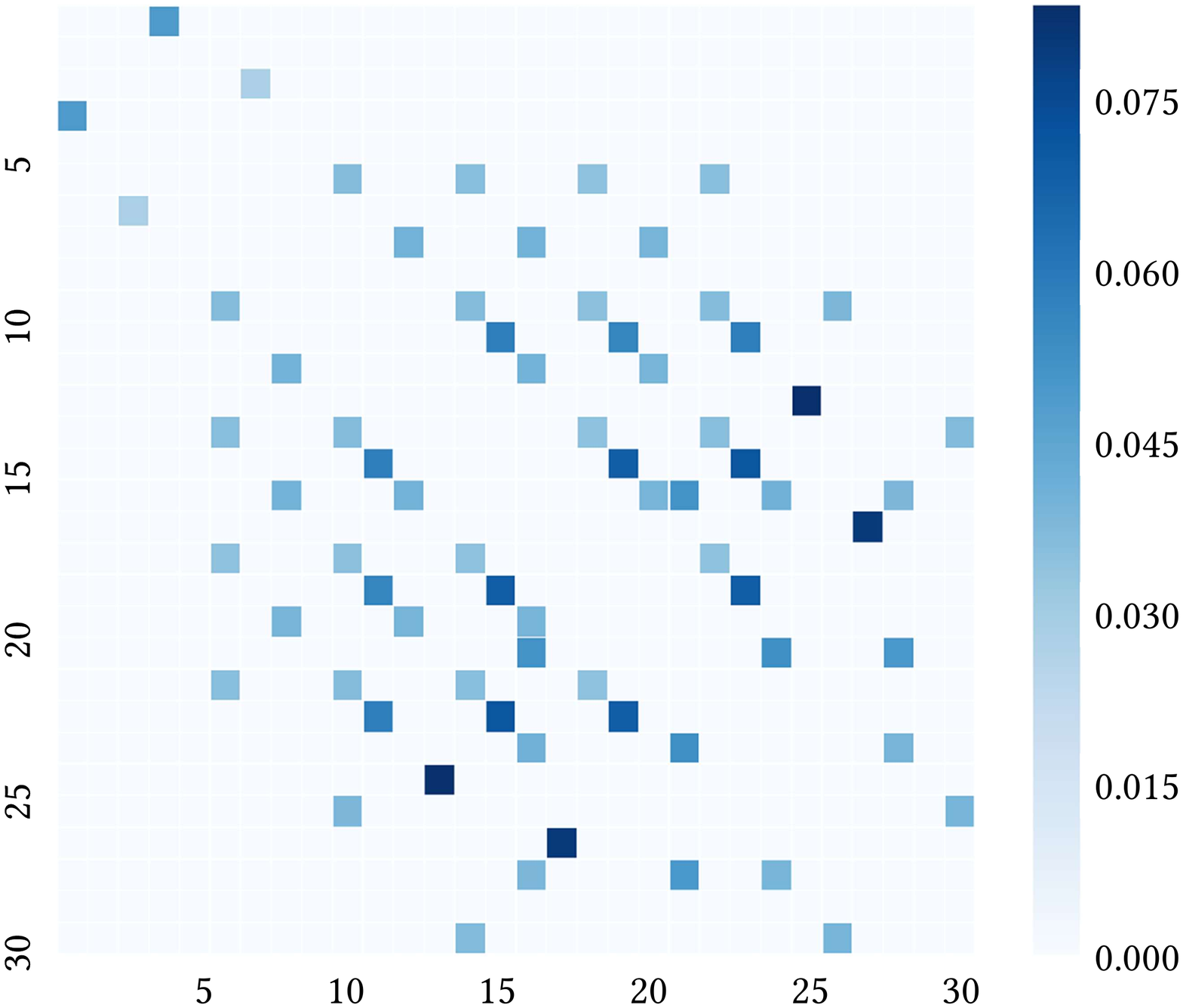}}
    \subfigure[Epoch 1]{\includegraphics[width=0.19\textwidth]{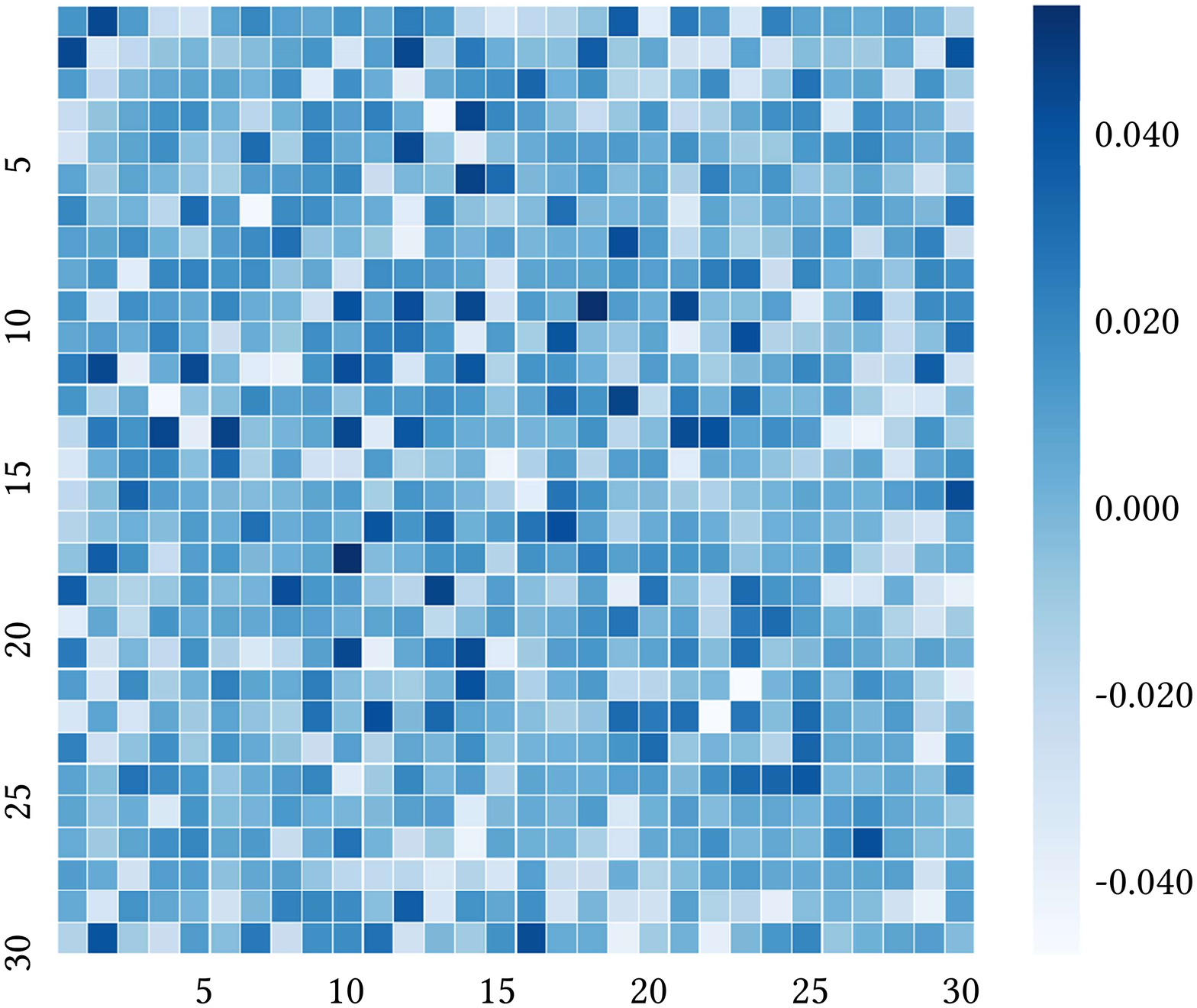}}
    \subfigure[Epoch 5]{\includegraphics[width=0.19\textwidth]{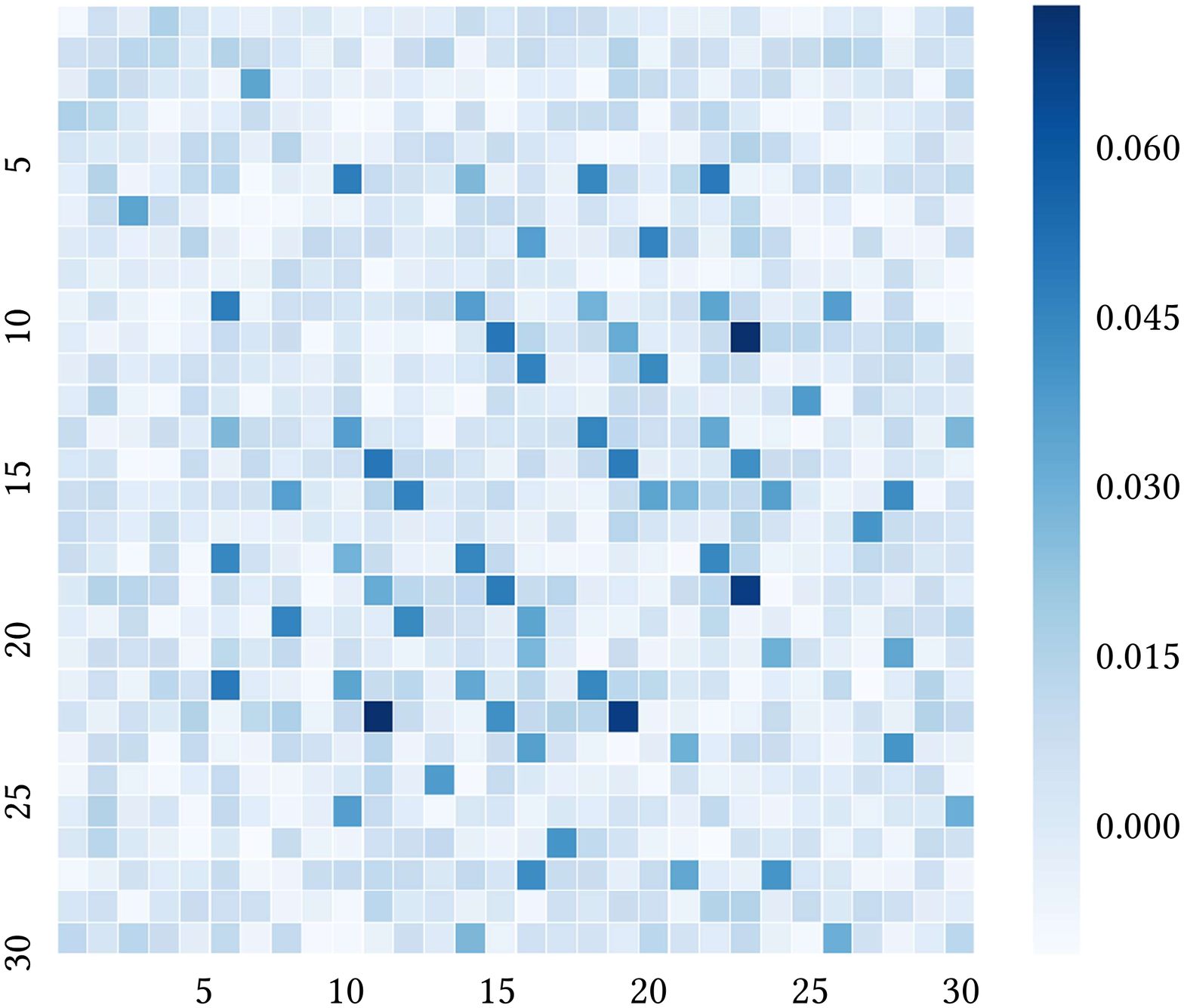}}
    \subfigure[Epoch 15]{\includegraphics[width=0.19\textwidth]{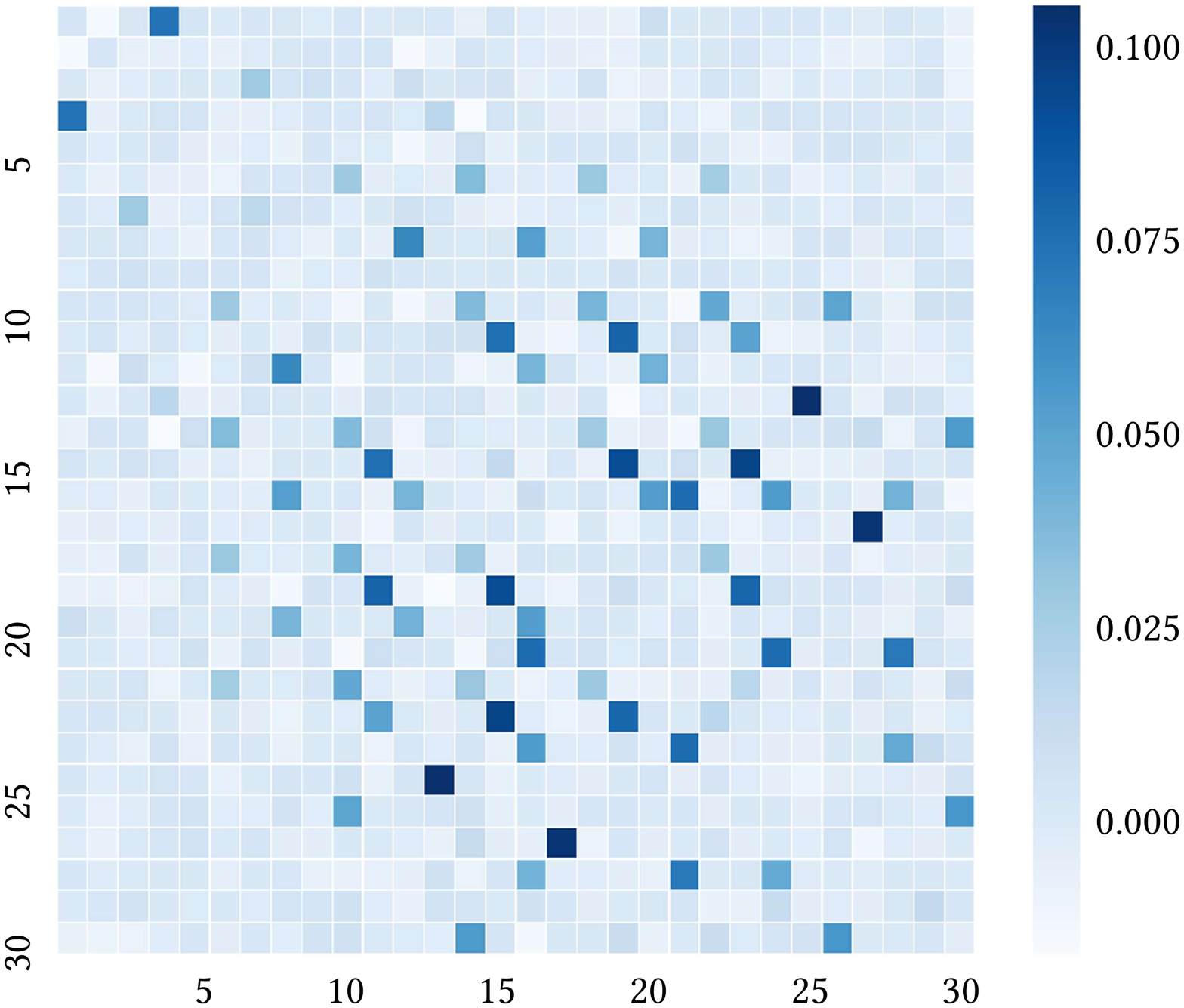}}
    \subfigure[Epoch 50]{\includegraphics[width=0.19\textwidth]{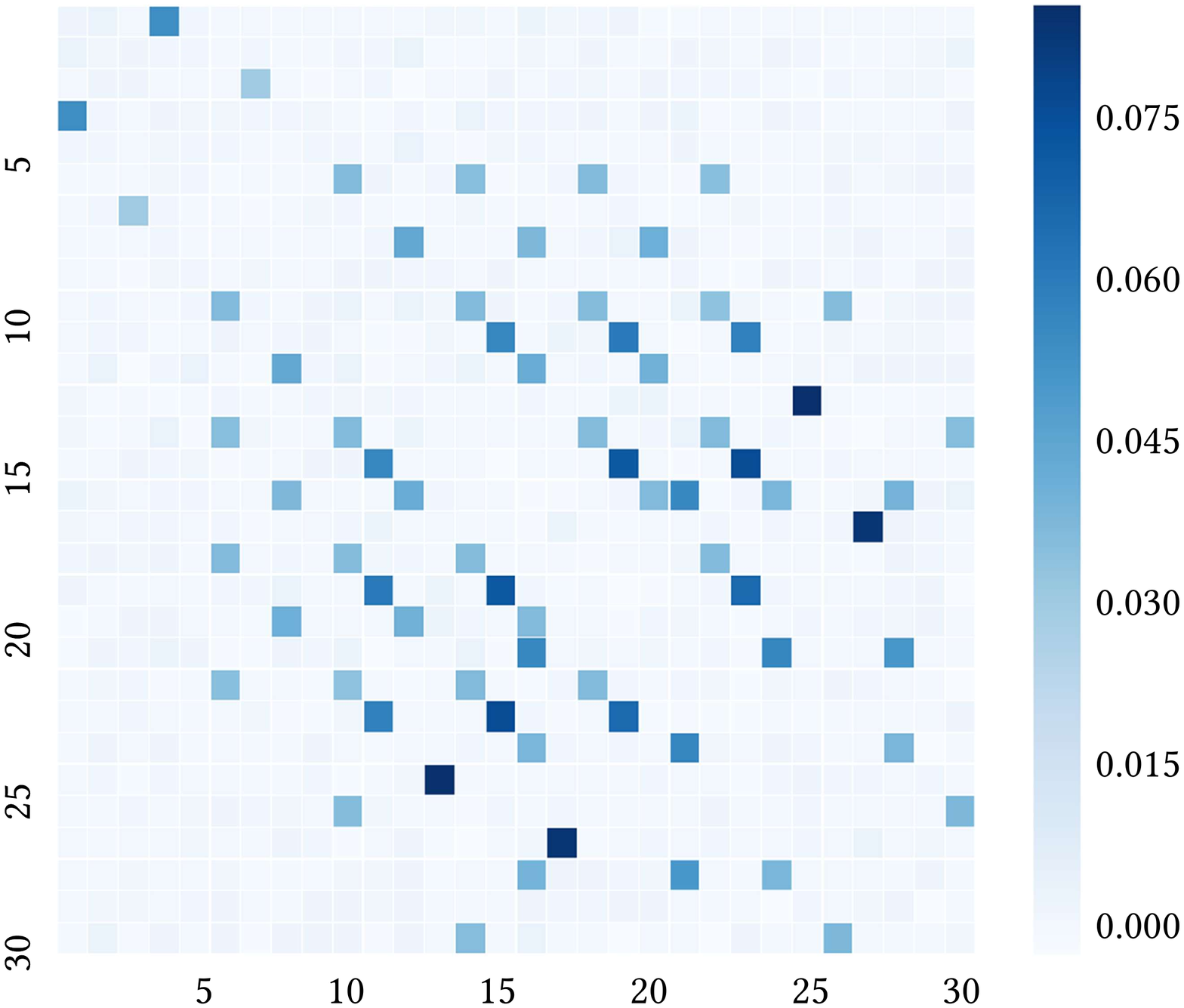}}
    \subfigure[Terror Attack: GT]{\includegraphics[width=0.19\textwidth]{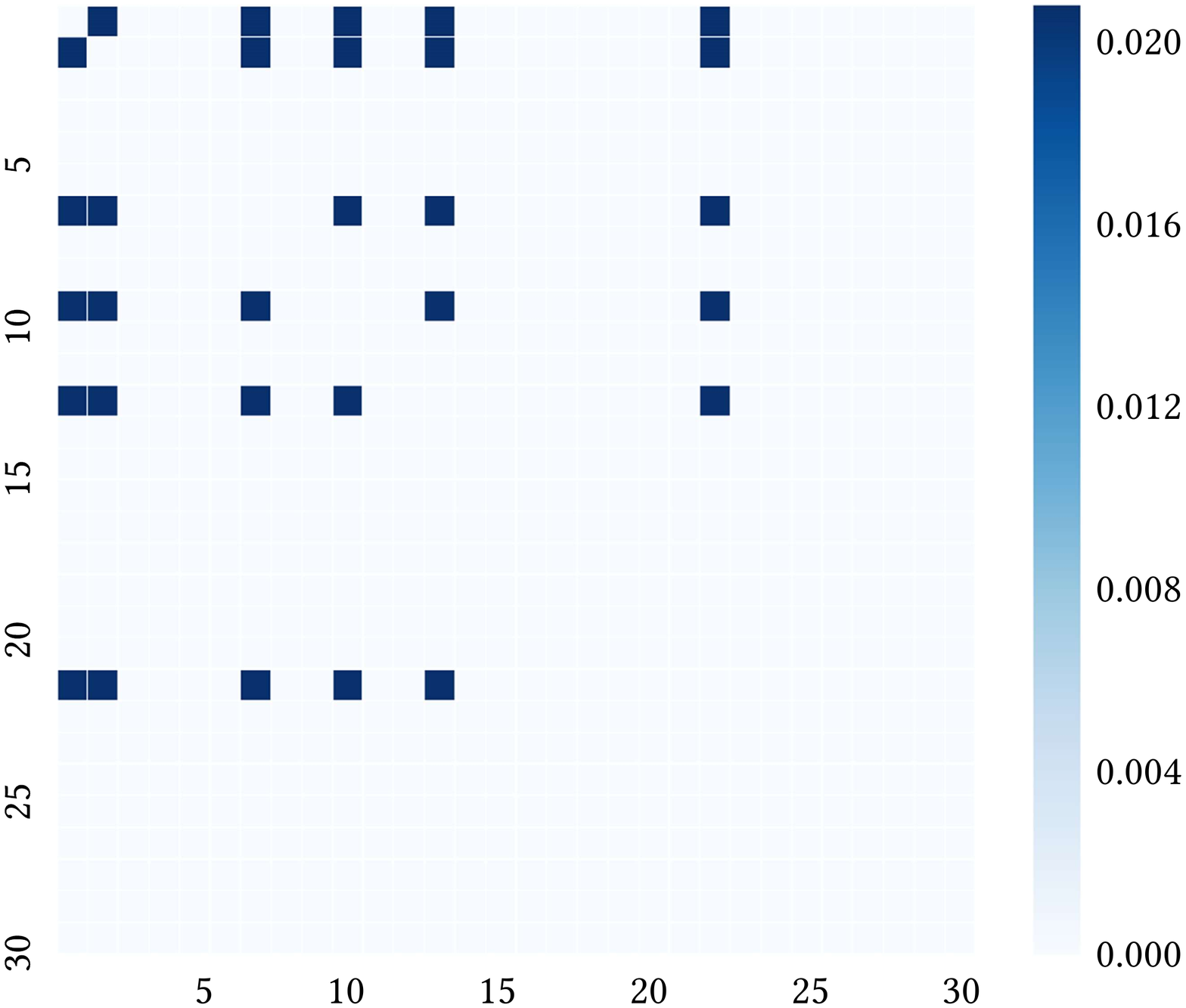}}
    \subfigure[Epoch 1]{\includegraphics[width=0.19\textwidth]{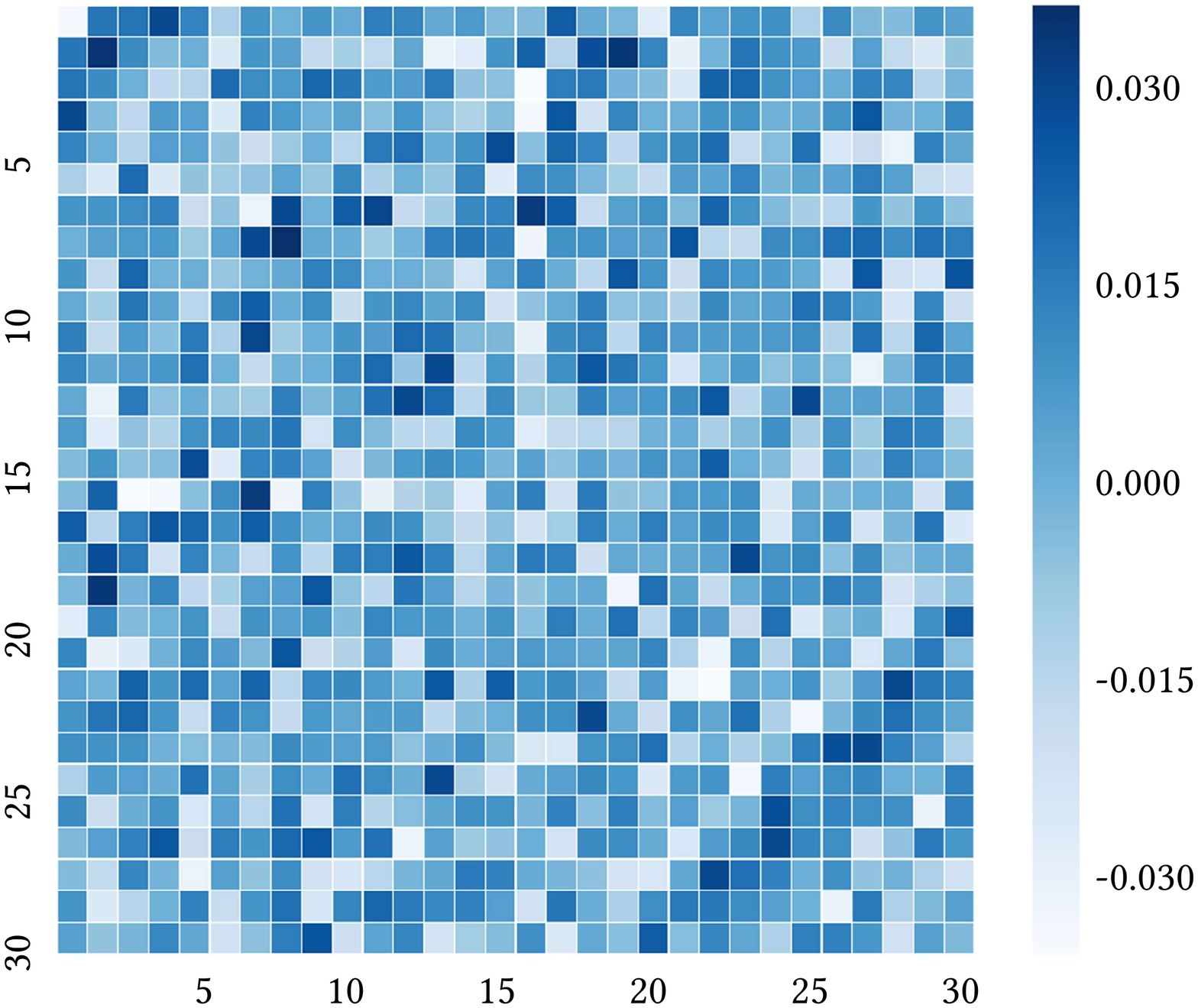}}
    \subfigure[Epoch 5]{\includegraphics[width=0.19\textwidth]{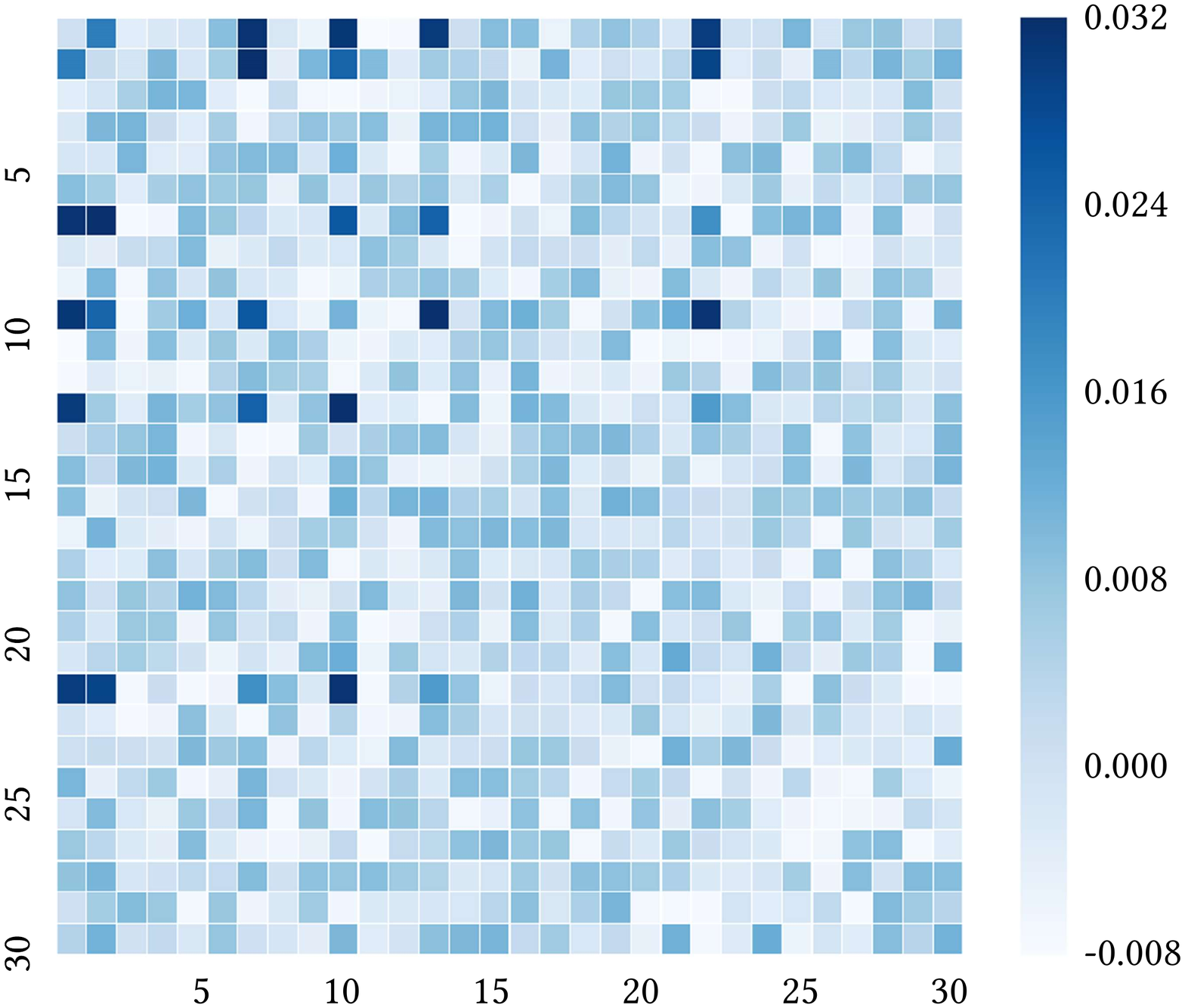}}
    \subfigure[Epoch 15]{\includegraphics[width=0.19\textwidth]{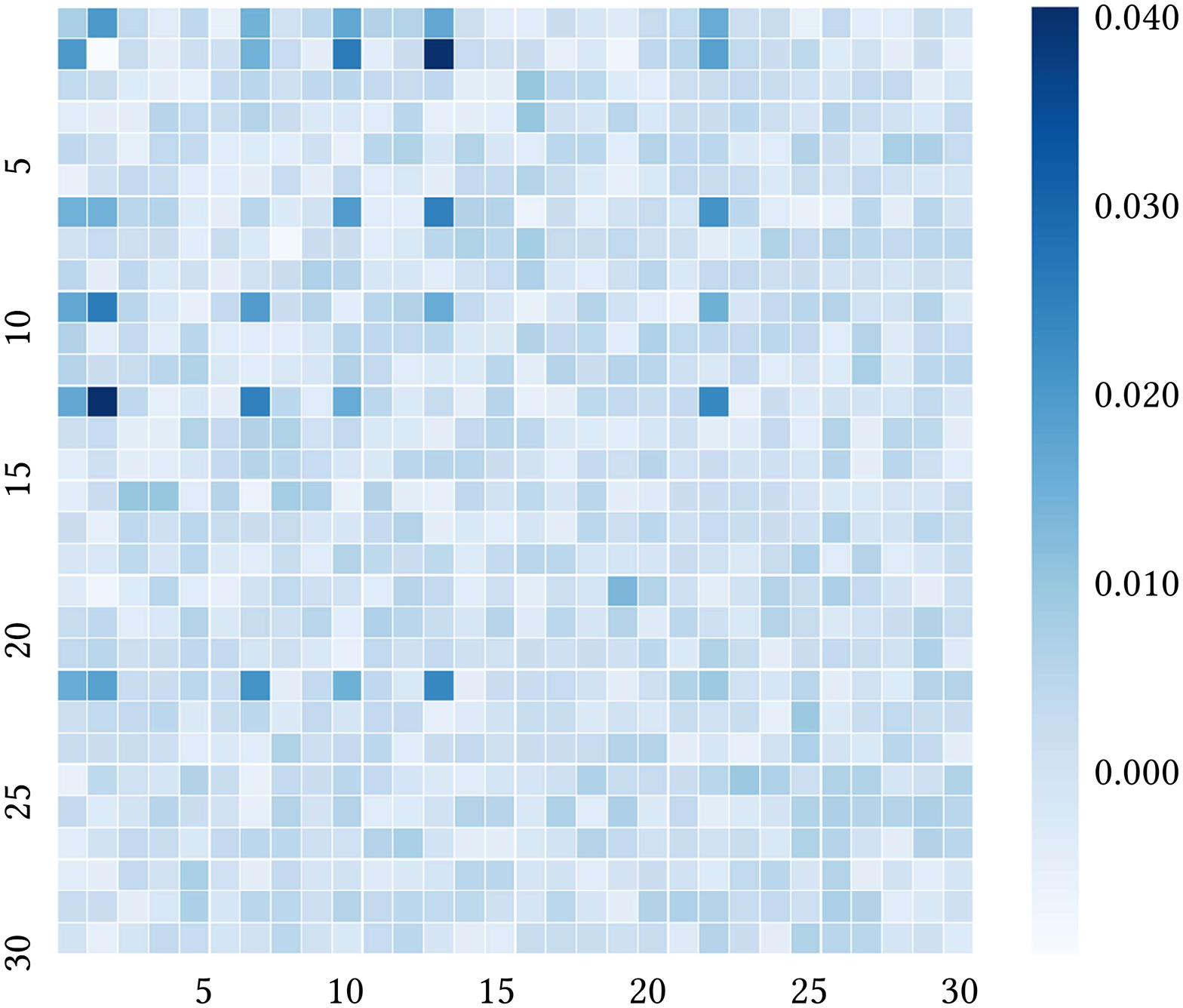}}
    \subfigure[Epoch 50]{\includegraphics[width=0.19\textwidth]{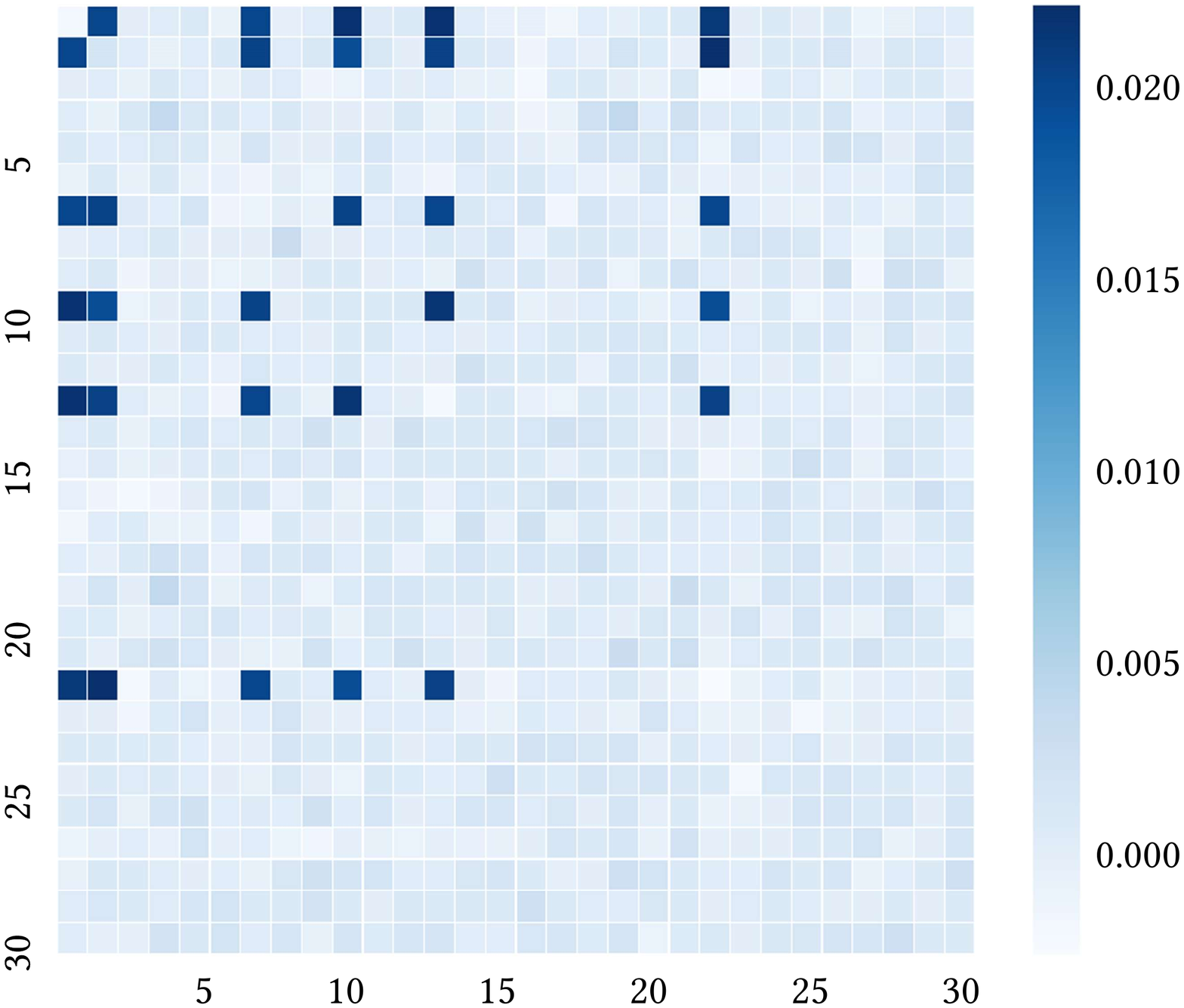}}
    \subfigure[Cora: GT]{\includegraphics[width=0.19\textwidth]{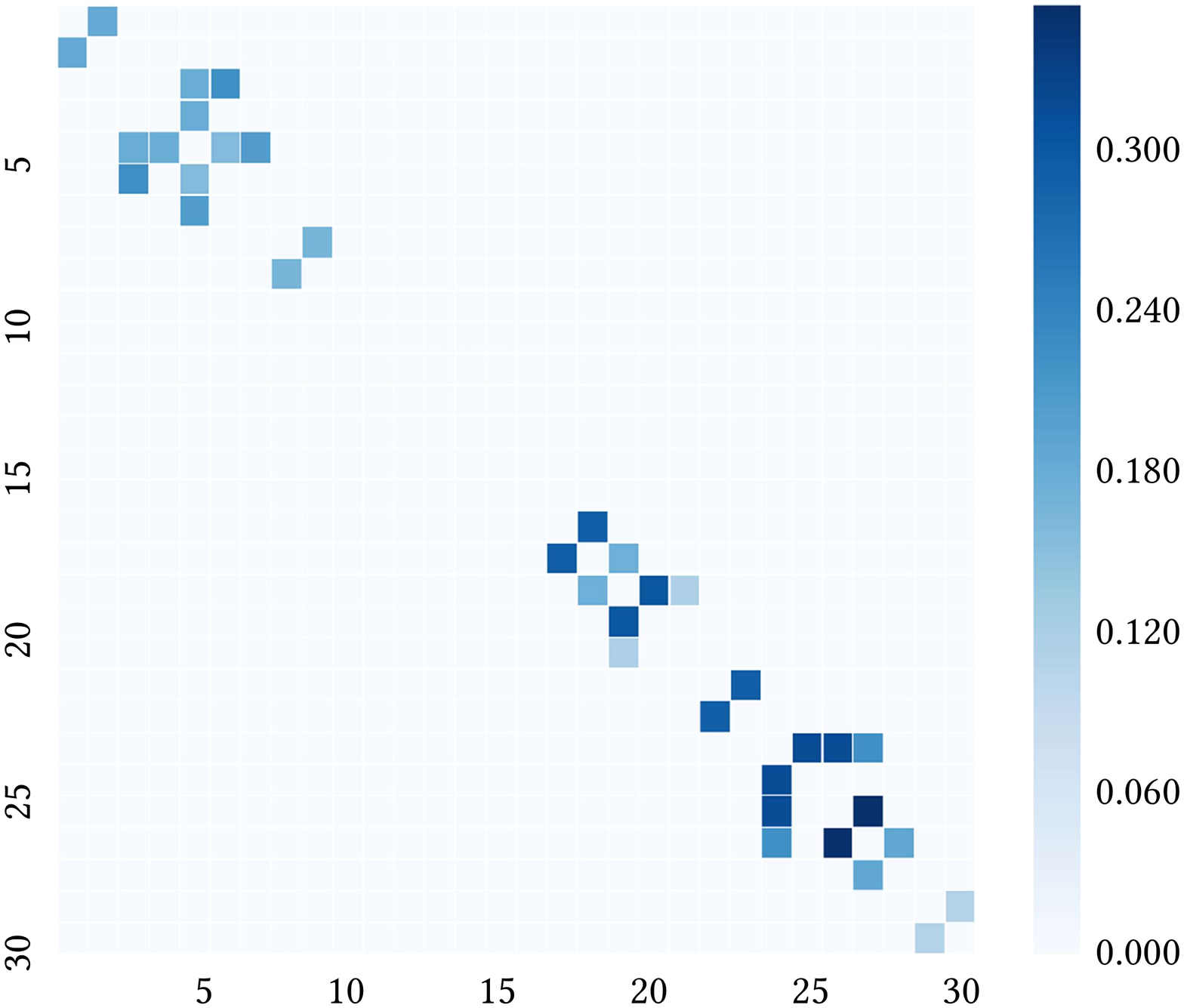}}
    \subfigure[Epoch 1]{\includegraphics[width=0.19\textwidth]{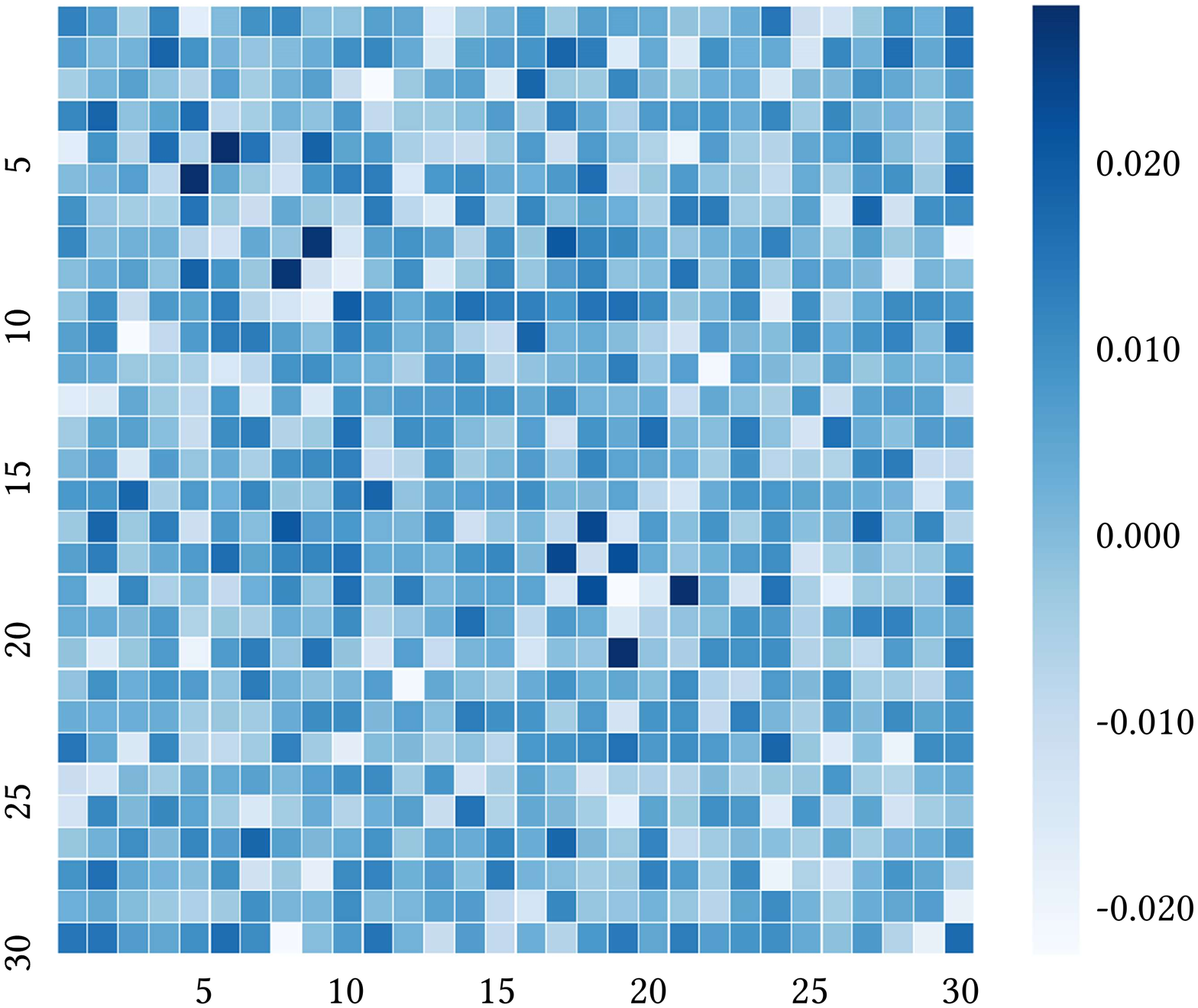}}
    \subfigure[Epoch 5]{\includegraphics[width=0.19\textwidth]{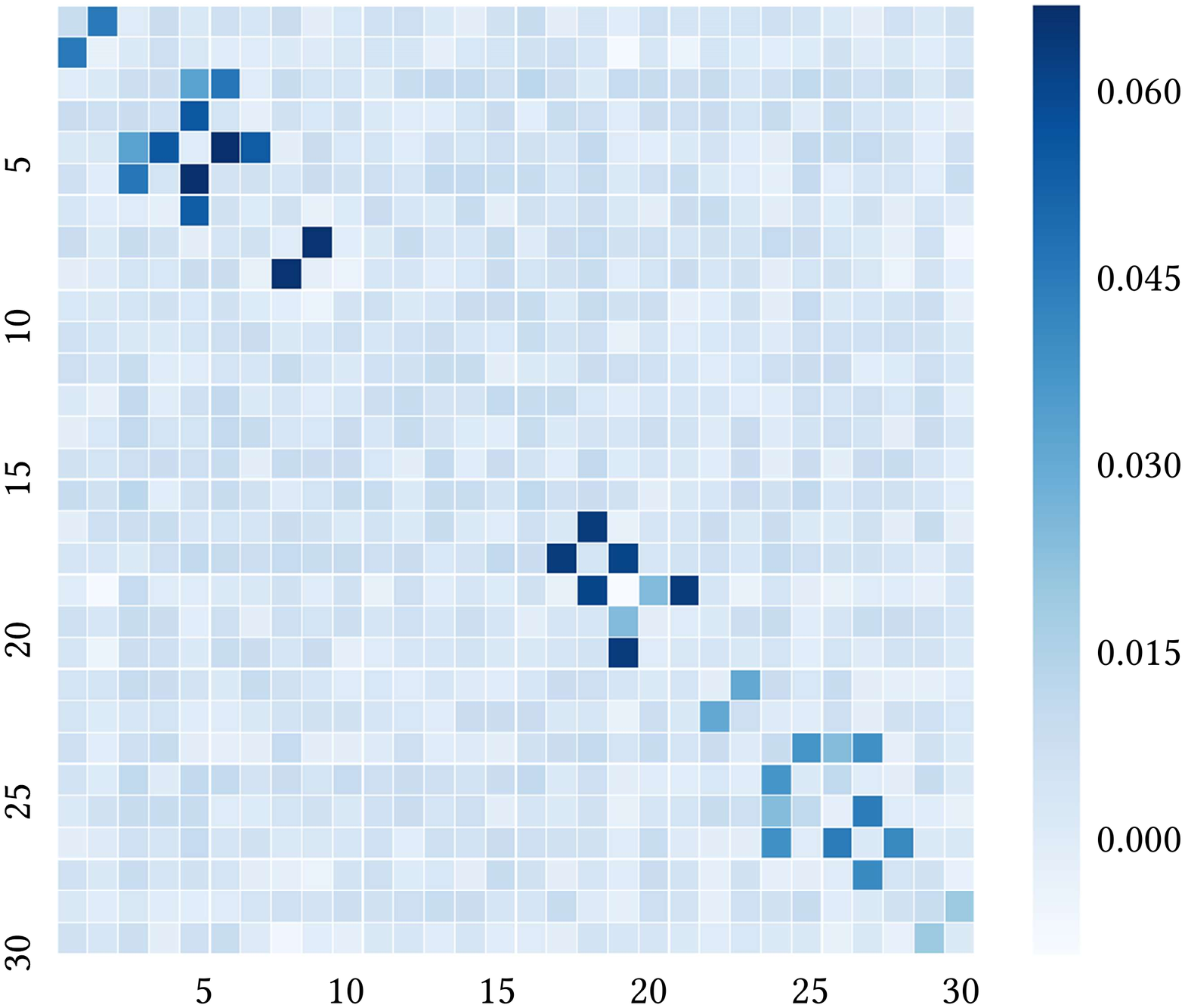}}
    \subfigure[Epoch 15]{\includegraphics[width=0.19\textwidth]{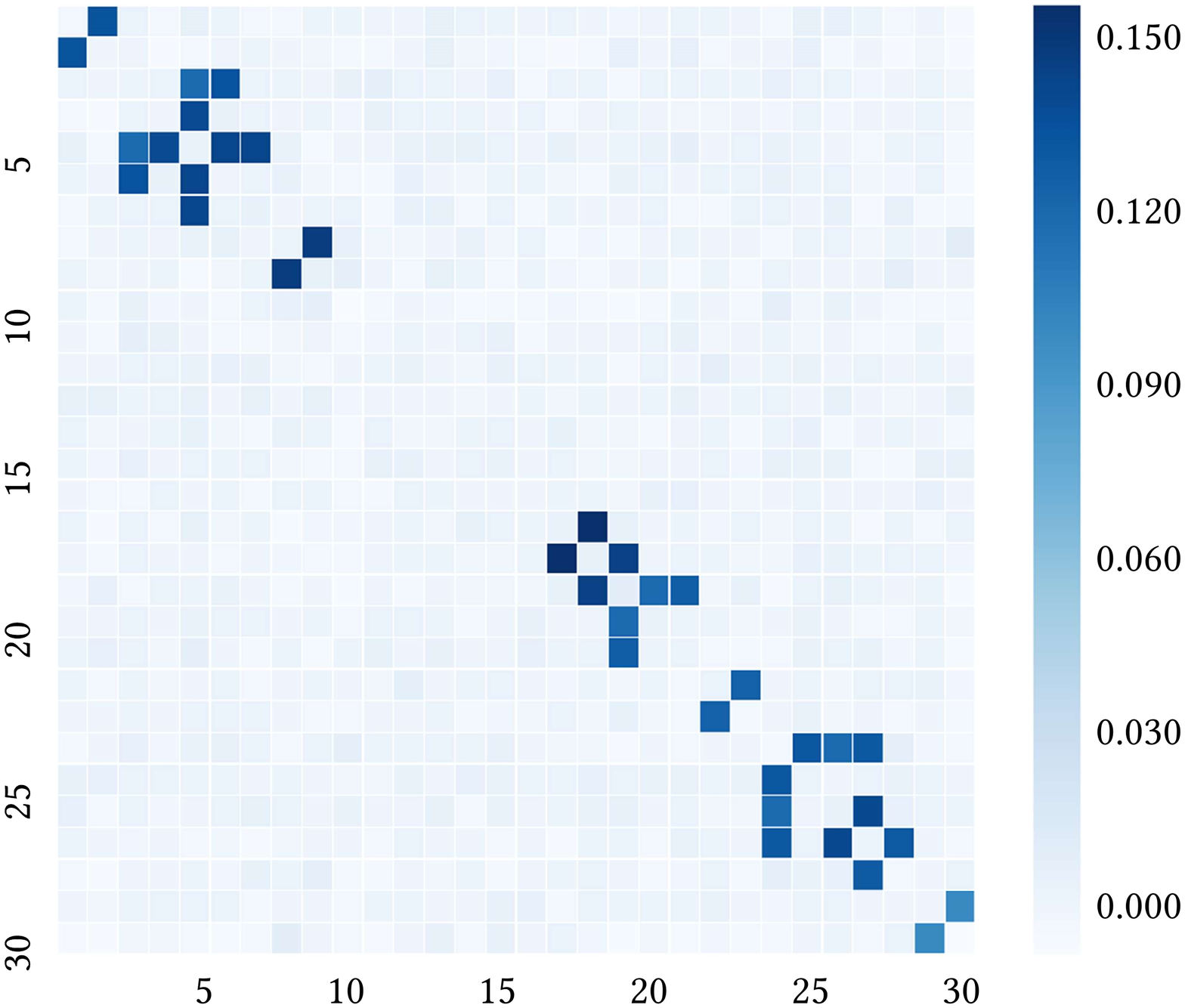}}
    \subfigure[Epoch 50]{\includegraphics[width=0.19\textwidth]{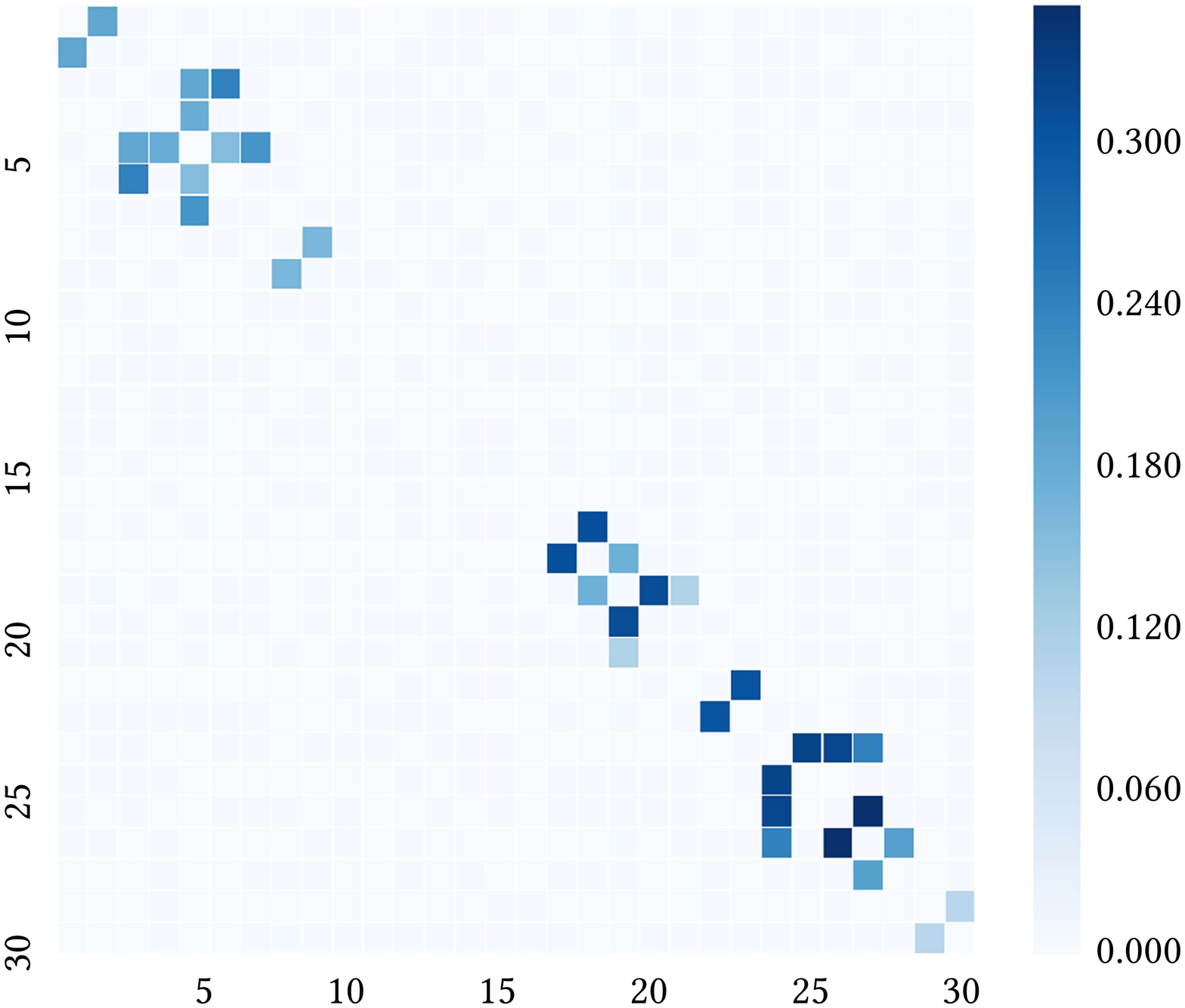}}
    \vspace{-0.1in}
    \caption{\textbf{The learned graph adjacency matrix (submatrix with size $\mathbf{30 \times 30}$).} The columns from the left to the right are the ground truth (GT) adjacency matrix, the learned adjacency matrix after the $1^\text{st}$, $5^\text{th}$, $15^\text{th}$ and $50^\text{th}$ epoch, respectively. The rows from the top to the bottom are for the TerroristsRel, Terror Attack, and Cora datasets, respectively.}
    \label{fig:visual}
\end{figure*}

%% file: 05_results.tex
In order to evaluate the performance of GLNN, we carry out extensive experiments on semi-supervised node classification tasks, and provide comparison with the state-of-the-art methods. Further, we evaluate the robustness of GLNN under different label rates. Finally, we visualize the learned graph adjacency matrix from GLNN and provide analysis.

\subsection{Datasets}

We consider two social networks: TerroristRel and Terror Attack \cite{zhao2006entity}, and three citation networks: Citeseer, Cora, and Pubmed \cite{sen2008collective}. The five datasets are available at https://linqs.soe.ucsc.edu/data. We represent the relationship between terrorists and citation links as undirected edges, and construct a binary and symmetric matrix as introduced in \cite{kipf2016semi} as the ground truth adjacency matrix. 



\textbf{TerroristRel.} This dataset contains information about terrorists and their relationships. We treat each terrorist as a node, and describe the relationship between each pair of terrorists via an edge in the graph. Further, we employ 160 nodes as labeled data to train our model, and leave the rest of 150 nodes for testing.

\begin{table*}[ht]
\caption{Results of node classification in terms of accuracy (\%). The asterisk indicates that we have downsampled the dataset.}
\centering
\begin{tabular}{rccccc}
\hline
\textbf{Method} & \textbf{TerroristsRel} & \textbf{Terror Attack} & \textbf{Citeseer} & \textbf{Cora} & \textbf{Pubmed*} \\ \hline
Planetoid \cite{yang2016revisiting} & - & - & 64.7 & 75.7 & 74.5 \\
ChebNet \cite{defferrard2016convolutional} & 60.1 & 66.5 & 69.6 & 81.2 & 71.6 \\
MoNet \cite{monti2017geometric} & 59.2 & 62.4 & - & 81.7 & 73.5 \\
LoNGAE \cite{tran2018learning} & 63.5 & 66.0 & 71.8 & 79.0 & 74.7  \\
GAT \cite{velivckovic2018graph} & 63.5 & 65.5 & 71.0 & 82.3 & 72.3  \\
DGI \cite{velickovic2019deep} & 52.7 & 61.5 & 71.8 & 82.3 & 74.6 \\
GWNN \cite{xu2018graph}  & 65.5 & 65.0 & 71.7 & 82.8 & 72.4  \\ \hline
GCN \cite{kipf2016semi} & 62.2 & 68.0 & 70.3 & 81.5 & 73.8 \\
\textbf{GLNN} & \textbf{70.9} & \textbf{76.5} & \textbf{72.4} & \textbf{83.4} & \textbf{76.7} \\ \hline
\end{tabular}
\label{tab:results}
\end{table*}

\textbf{Terror Attack.} This dataset consists of 1,293 terrorist attacks, each of which is assigned one of 6 labels to indicate the attack type. We treat terror attacks as nodes, and describe their relationship via edges. We deploy 120 nodes for training and leave 200 nodes for testing.

\textbf{Citation networks.} We consider three widely used citation network datasets: Citeseer, Cora, and Pubmed. In these datasets, nodes are documents and edges are citation links. In particular, we sample 3,000 nodes to evaluate our model for the Pubmed dataset. In addition, we select 60 nodes as labeled data to train each model and test the performance on 1,000 nodes. For the Citeseer and Cora dataset, we follow the experimental setup in (Yang et al. \cite{yang2016revisiting}).

\vspace{-0.1in}
\subsection{Implementation Details}

We implemented the proposed model with the TensorFlow\footnote{https://tensorflow.google.cn} framework, and empolyed Adam \cite{kingma2014adam} to train the entire network with the learning rate 0.01. Before the first epoch of the training step in the graph learning layer, we generate a random matrix that follows a uniform distribution, which serves as the initial graph representation. This matrix is then optimized in the subsequent training epochs. In the graph learning loss function, we set 
$\lambda_1 = 0.1$, $\lambda_3 = 0.1$, and $\lambda_4 = 0.001$. Note that $\lambda_2$ is set to $0$ with the symmetrization operation introduced in (\ref{eq:symmetrization}). In the overall loss function (\ref{eq:overall_loss}), we set $\alpha = 10$.
The settings of $\lambda_0$ are adaptive to each dataset, as will be discussed in the subsequent \textit{GLR Analysis} subsection.
In the graph convolutional layers, we follow the same experimental settings in \cite{kipf2016semi}, i.e., two graph convolutional layers with 16 hidden units.

\subsection{Node Classification Results}

Table~\ref{tab:results} lists the comparison results on the five datasets. Reported numbers denote the classification accuracy in percentage, and the best results are marked in bold. Note that the underlying graphs of TerroristsRel and Terror Attack datasets are disconnected, so the Planetoid \cite{yang2016revisiting} cannot be executed properly.

First, we see that our GLNN outperforms the baseline method GCN (Kipf et al. \cite{kipf2016semi}) on all the datasets. This validates the effectiveness of the proposed graph learning for semi-supervised classification. In addition, GLNN outperforms the recently proposed network GWNN \cite{xu2018graph}, which indicates the benefit of GLNN on graph data representation and learning. Further, GLNN achieves better performance than other semi-supervised learning methods, which demonstrates the superiority of our model on conducting semi-supervised classification tasks on graph data.

\begin{table}
\caption{Results of node classification in terms of accuracy with different $\lambda_0$.}
\centering
\begin{tabular}{c|p{0.5cm}<{\centering}p{0.63cm}<{\centering}p{0.63cm}<{\centering}p{0.63cm}<{\centering}p{0.63cm}<{\centering}p{0.5cm}<{\centering}} \hline
\multirow{2}{*}{Dataset~} & \multicolumn{5}{c}{Values of $\lambda_0$}  \\ \cline{2-7}
                          & 1.0 & $10^{-1}$ & $10^{-2}$ & $10^{-3}$ & $10^{-4}$ & 0.0 \\ \hline
TerroristsRel & 52.0 & \textbf{70.9} & \textbf{70.9} & 63.5 & 63.5 & 62.8 \\
Terror Attack & \textbf{76.5} & \textbf{76.5} & 67.5 & 51.5 & 57.5 & 57.0 \\
Citeseer & \textbf{72.4} & 71.2 & 72.2 & 70.4 & 70.4 & 69.8 \\
Cora & 76.7 & 80.0 & \textbf{83.4} & 82.0  & 82.1 & 82.1 \\
Pubmed* & \textbf{76.7} & 74.8 & 73.0 & 73.0 & 72.9 & 72.9 \\ \hline
\end{tabular}
\label{tab:para_analysis}
\end{table}

\subsection{Robustness Test}

\begin{figure}[htbp]
    \centering
    \subfigure[Citeseer dataset.]{\includegraphics[width=0.45\textwidth]{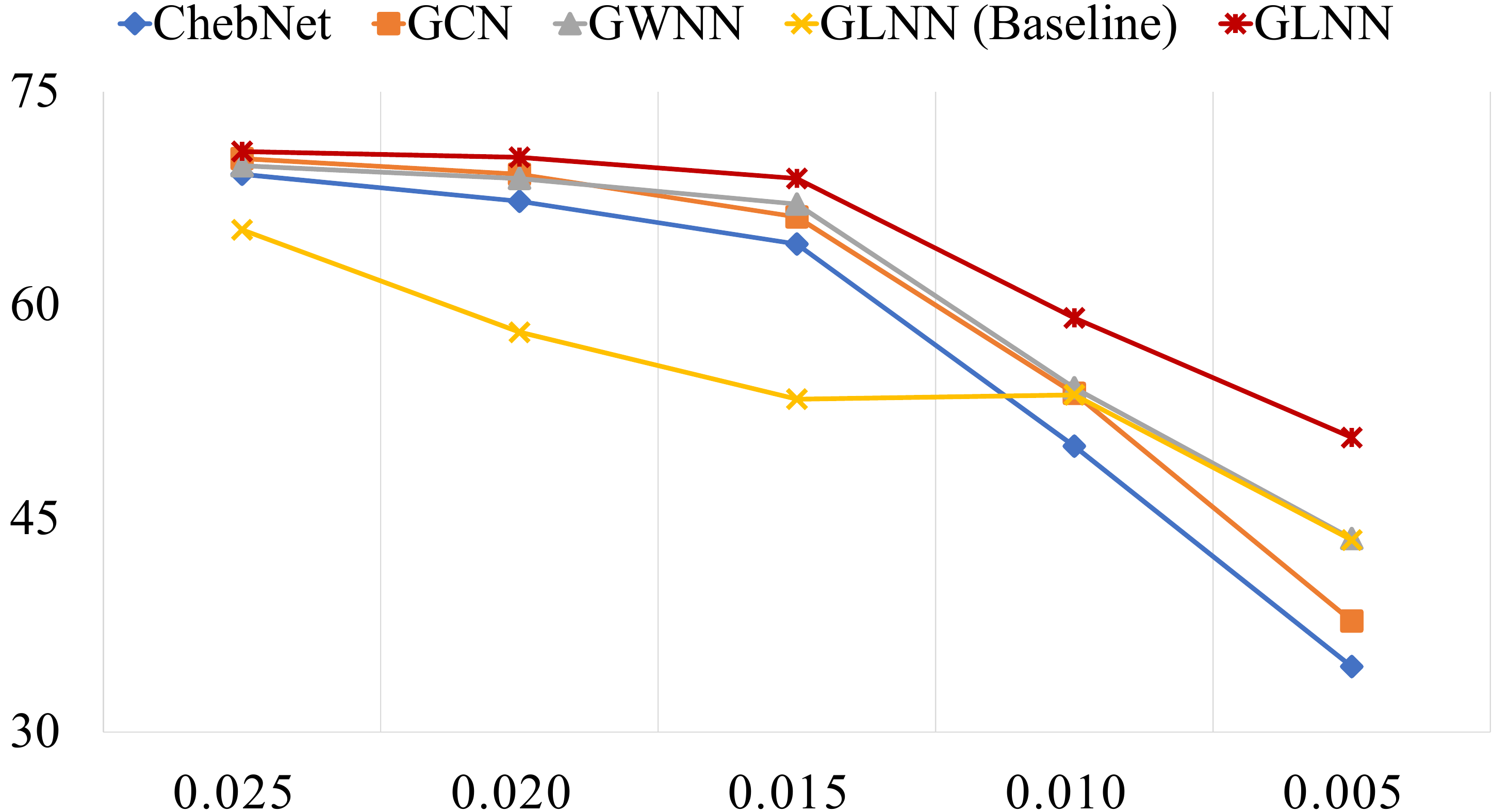}}
    \hspace{0.3in}
 	\subfigure[Cora dataset.]{\includegraphics[width=0.45\textwidth]{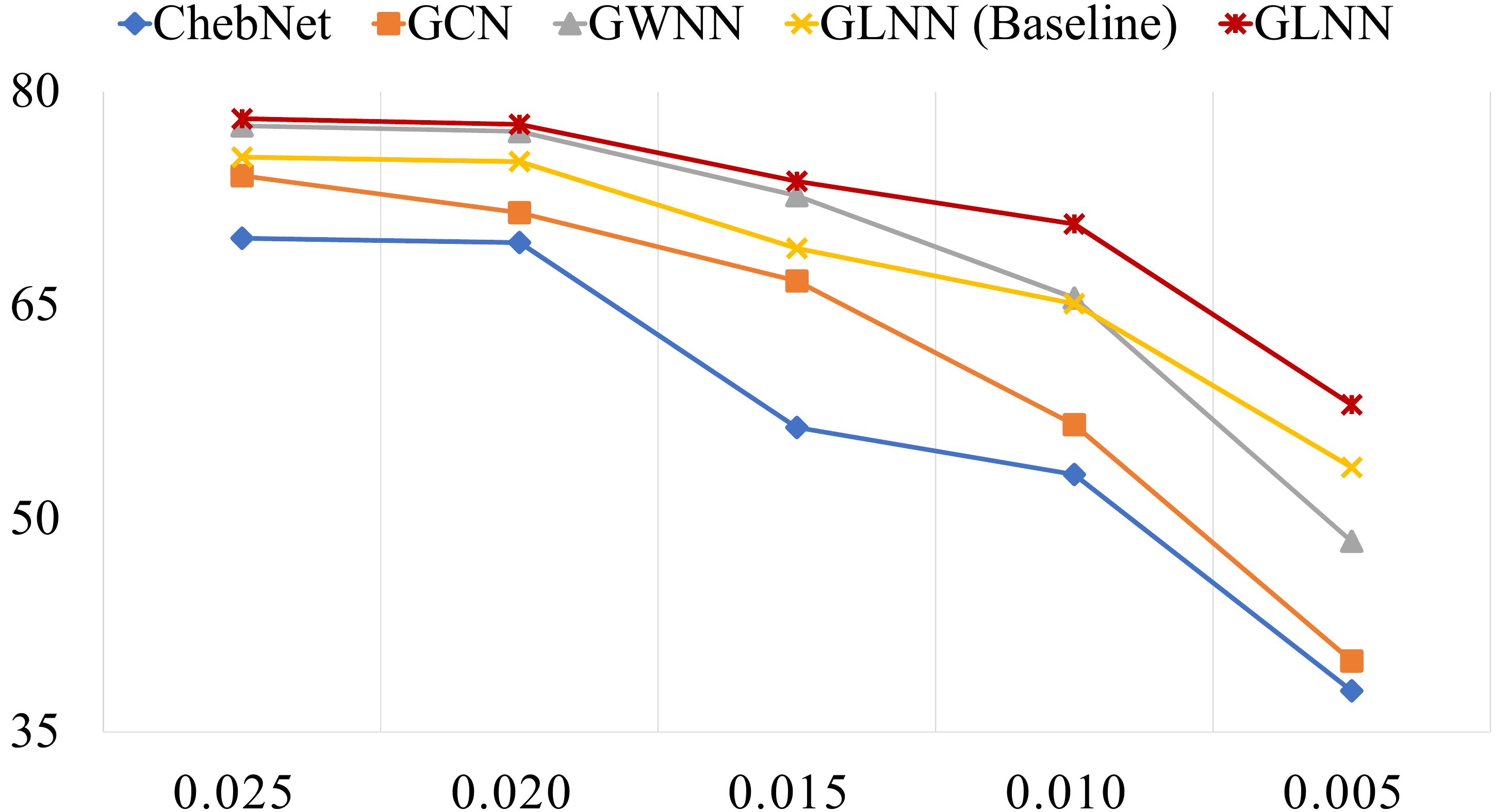}}
 	\vspace{-0.1in}
    \caption{Classification accuracy comparison with different label rates.}
    \label{fig:robustness}
\end{figure}

We test the robustness of our model on citation networks under low label rates. We adopt five different label rates in \{0.005, 0.010, 0.015, 0.020, 0.025\} to train our model and three representative baseline methods: ChebNet \cite{defferrard2016convolutional}, GCN \cite{kipf2016semi} and GWNN \cite{xu2018graph}. Fig.~\ref{fig:robustness} presents the classification accuracy under five different label rates for GCN, ChebNet, GWNN, and our method on the Citeseer and Cora datasets respectively.
We see that while the performance of the other methods drops quickly with decreasing label rate, the proposed GLNN is much more robust, with the classification accuracy of $51.8\%$ and $58.0\%$ on the Citeseer and Cora datasets even when the label rate is lower than $0.01$.

Further, to validate the necessity and effectiveness of our proposed GLR loss, we also compare with a \textbf{Baseline} scheme of our model, where the GLR loss term $\mathcal{L}_\text{GLR}$ is removed from (\ref{eq:loss_gl}).
As shown in Fig.~\ref{fig:robustness}, we outperform the \textbf{Baseline} scheme by a large margin, which validates the superiority of our proposed GLR loss function. To sum up, the robustness of our proposed GLNN gives credits to the structure-adaptive regularization---\textit{Graph Laplacian Regularizer}---in the loss function. This prior is critical in practical applications, since graph data in real life are often unlabeled or have few accurate labels. 

\subsection{GLR Analysis}

We evaluate the contribution of GLR via different assignments of the GLR weighting parameter $\lambda_0$. Table~\ref{tab:para_analysis} lists the classification results of five datasets with $\lambda_0 \in \{1.0, 10^{-1}, 10^{-2}, 10^{-3}, 10^{-4}, 0.0\}$. 
We see that when $\lambda_0$ is set to a small value, GLNN can produce reasonable results. 
The performance keeps improving with increasing value of $\lambda_0$ in general, finally achieving state-of-the-art performance.  
This shows that the GLR loss term with higher weights makes significant contribution to classification results, which demonstrates the superiority of GLR-based graph learning.

\subsection{Visualization and Analysis}

We also provide some visualization results of the learned graph adjacency matrix obtained from GLNN in Fig.~\ref{fig:visual}. From the left to right columns, we show the ground truth adjacency matrix, the learned adjacency matrix after the $1^\text{st}$, $5^\text{th}$, $15^\text{th}$ and $50^\text{th}$ epoch, respectively. Note that, the adjacency matrix at the $1^\text{st}$ epoch is an initialized random matrix that follows a uniform distribution. In order to ensure clear visualization, we select a sub-matrix with size $30 \times 30$ from the entire adjacency matrix. 

We see that for all the three datasets, as the epoch increases, the learned graph adjacency matrix becomes more and more similar with the ground truth one. At the $50^\text{th}$ epoch, the learned adjacency matrix is almost the same as the ground truth while introducing , thus validating the effectiveness of the proposed graph learning. Also, the learned adjacency matrix becomes sparser as the epoch increases, due to the sparsity constraint in the proposed graph learning loss.    

%% file: 06_conclusion.tex

In this paper, we propose a novel Graph Learning Neural Network (GLNN), aiming to learn an optimal graph from both data and tasks for robust semi-supervised classification. 
Our GLNN model exploits structure-adaptive graph learning based on graph Laplacian regularizer (GLR), which enforces the graph to capture pairwise similarities within data. 
We also pose the sparsity constraint and properties of a valid adjacency matrix, which formulates the optimization of the underlying graph along with GLR. 
The optimization objective is then integrated into the loss function of a GCNN.  
Experimental results on widely adopted citation network datasets and social network datasets validate the effectiveness and robustness of the proposed GLNN especially at low label rates. 
In the future, we will extend our GLNN to other applications such as 3D point cloud classification and segmentation, where there is no ground truth graph available for training data.